\documentclass[lettersize,journal]{IEEEtran}
\usepackage{amsmath,amsfonts}
\usepackage{algorithmic}
\usepackage{array}
\usepackage[caption=false,font=normalsize,labelfont=sf,textfont=sf]{subfig}
\usepackage{textcomp}
\usepackage{stfloats}
\usepackage{url}
\usepackage{verbatim}
\usepackage[numbers]{natbib}
\usepackage{graphicx}

\def\BibTeX{{\rm B\kern-.05em{\sc i\kern-.025em b}\kern-.08em
    T\kern-.1667em\lower.7ex\hbox{E}\kern-.125emX}}
\usepackage{balance}
\usepackage{multirow}
\usepackage[table,xcdraw]{xcolor}
\newcommand{\cmark}{\color{black}{\ding{51}}}
\newcommand{\xmark}{\color{black}{\ding{55}}}
\usepackage{pifont}
\usepackage{algorithm}

\usepackage{times}
\usepackage{epsfig}
\usepackage{amssymb}
\usepackage{booktabs}
\usepackage{commath}
\usepackage{mathtools}
\usepackage{dsfont}
\usepackage{lipsum}
\usepackage{bbm}
\usepackage{makecell}

\usepackage{fancyhdr}
\usepackage{anyfontsize}
\usepackage{hyperref}

\definecolor{Gray}{gray}{0.9}
\definecolor{darkmagenta}{RGB}{127,0,127}
 
\definecolor{coolblack}{rgb}{0.0, 0.23, 0.64}
\newcommand{\jnkc}[1]{\textcolor{coolblack}{#1}} 
\newcommand{\imp}[1]{\textcolor{darkmagenta}{#1}} 

\begin{document}
\title{Robust Few-shot Learning Without Using any Adversarial Samples}

\author{Gaurav Kumar Nayak,~\IEEEmembership{Graduate Student Member, IEEE}, Ruchit Rawal, Inder Khatri and Anirban~Chakraborty,~\IEEEmembership{Member, IEEE}
\IEEEcompsocitemizethanks{
\IEEEcompsocthanksitem G. K. Nayak, R. Rawal, I. Khatri and A. Chakraborty are with the Department of Computational and Data Sciences, Indian Institute of Science, Bangalore, India, 560012.
}
\thanks{For all correspondence: Anirban Chakraborty (anirban@iisc.ac.in)}
}






\markboth{Journal of \LaTeX\ Class Files,~Vol.~18, No.~9, September~2020}%
{How to Use the IEEEtran \LaTeX \ Templates}

\maketitle

\begin{abstract}
   The high cost of acquiring and annotating samples has made the `few-shot' learning problem of prime importance. Existing works mainly focus on improving performance on clean data and overlook robustness concerns on the data perturbed with adversarial noise. Recently, a few efforts have been made to combine the few-shot problem with the robustness objective using sophisticated Meta-Learning techniques. These methods rely on the generation of adversarial samples in every episode of training, which further adds a computational burden. To avoid such time-consuming and complicated procedures, we propose a simple but effective alternative that does not require any adversarial samples. Inspired by the cognitive decision-making process in humans, we enforce high-level feature matching between the base class data and their corresponding low-frequency samples in the pretraining stage via self distillation. The model is then fine-tuned on the samples of novel classes where we additionally improve the discriminability of low-frequency query set features via cosine similarity. On a $1$-shot setting of the CIFAR-FS dataset, our method yields a massive improvement of $60.55\%$ \& $62.05\%$ in adversarial accuracy on the PGD and state-of-the-art Auto Attack, respectively, with a minor drop in clean accuracy compared to the baseline. Moreover, our method only takes $1.69\times$ of the standard training time while being $\approx 5\times$ faster than state-of-the-art adversarial meta-learning methods. The code is available at \href{https://github.com/vcl-iisc/robust-few-shot-learning}{https://github.com/vcl-iisc/robust-few-shot-learning}.
\end{abstract}


\section{Introduction}
\label{sec:intro}
\IEEEPARstart{T}{rained} deep models deployed in the real world are expected to yield reliable predictions. However, these models remain brittle even after attaining very high performance on the target task. In other words, even a slight perturbation in input image, which is human-imperceptible, changes the predictions on these models. Such perturbed images are often called `adversarial images'~\cite{szegedy2013intriguing,goodfellow2014explaining}. The perturbations added to the input image are called `adversarial noise', which are carefully crafted (via adversarial attack) with an objective to `fool' the trained network. Such attacks are a threat to trained deep neural networks and can lead to catastrophic consequences in applications such as self-driving cars~\cite{yang2020finding}, biometric authentication~\cite{fei2020adversarial} etc. Thus, there is an urgent need to defend the deep models against these attacks.

Adversarial training~\cite{madry2017towards} is one of the most effective and popular techniques for defense against adversarial attacks. However, its efficacy is heavily dependent on the amount of training data~\cite{carmon2019unlabeled,alayrac2019labels}. The adversarially trained models fail to resist adversarial attacks when scarce data or few training samples are present. Their performance gets worse, especially in the case of few-shot learning where the samples in the support set can be as low as one sample per class (also known as $1$-shot)~\cite{goldblum2020adversarially}. 

Popular methods of few-shot learning~\cite{snell2017prototypical,bertinetto2018metalearning,lee2019meta} only focus on improving performance on a clean query set. Their performance drops drastically when query samples are perturbed via adversarial attacks. Yin~\textit{et al.}~\cite{yin2018adversarial} showed that even the sophisticated meta-learning algorithms for few-shot learning such as Relation Networks~\cite{sung2018learning}, Matching networks~\cite{vinyals2016matching} and MAML~\cite{finn2017model} are susceptible to adversarial attacks. 
They proposed an adversarial meta learner, which improved the adversarial accuracy but only against weaker attacks. Their method depended on the MAML framework, and their overall robustness was still poor. 
Recently, 
Goldblum~\textit{et al.}~\cite{goldblum2020adversarially} 
combined adversarial training with meta-learning to overcome these problems. Importantly, their framework is agnostic to any specific meta-learning algorithm and showed better robustness. However, they require additional $n$+$1$ SGD steps in every episode of pretraining compared to standard meta-learning. In each episode, they need to generate adversarial samples through an optimization procedure which runs for $n$ iterations. Overall, these methods that are designed to include robustness in few-shot learning are computationally expensive and dependent on meta-learning, which involves complicated optimization. Also, these methods require constructing adversarial samples for every batch of training data, which itself needs several iterations of backpropagation, 
adding a further computational overhead. 

\begin{figure*}[t]
\centering
\centerline{\includegraphics[width=0.82\textwidth]{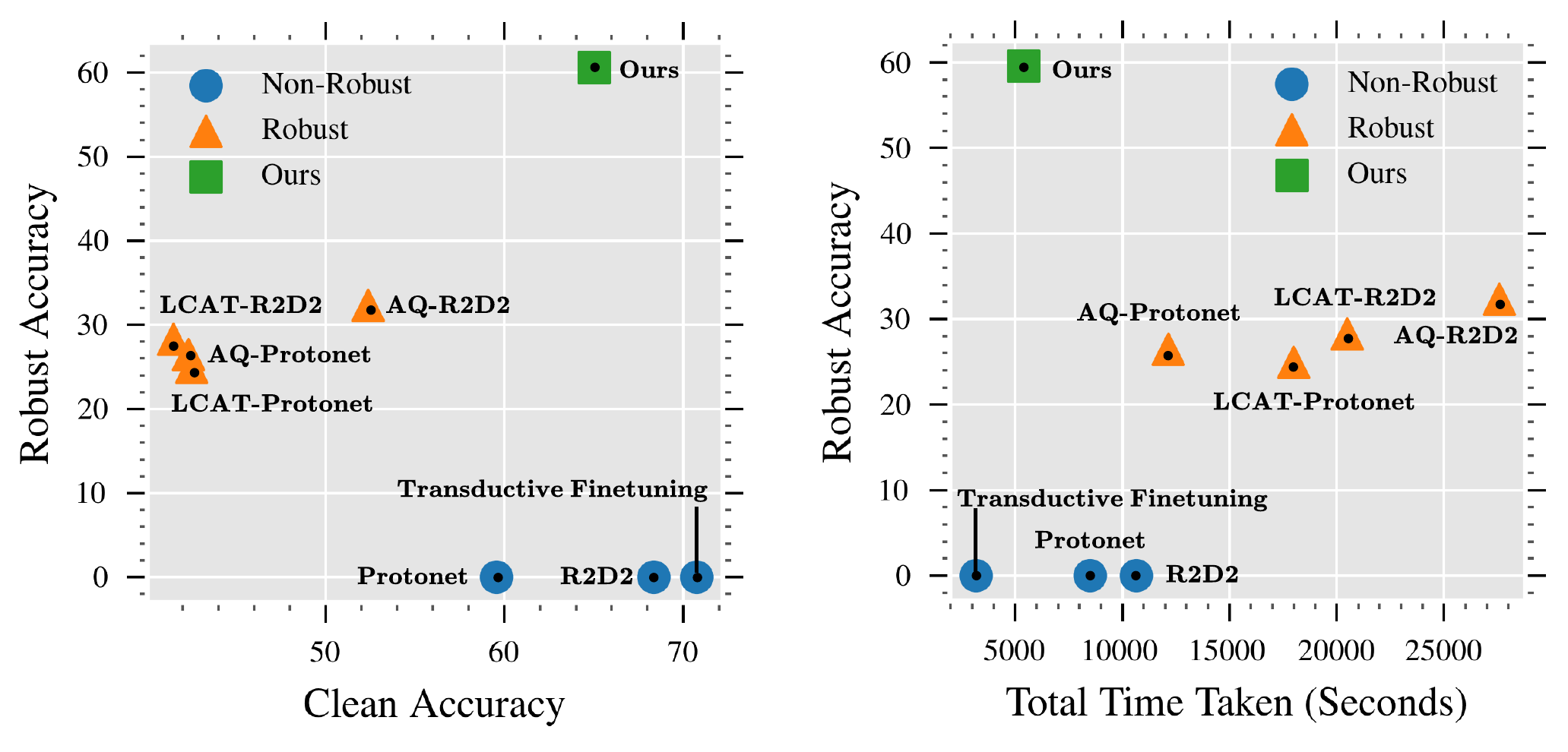}}
\caption{(Left) Robust acc. (on PGD) v/s Clean acc. and (Right) Robust acc. v/s Total Time taken (includes pretraining, finetuning and inference time for a given support and query set) for $1$-shot $5$-way setting on CIFAR-FS Dataset. Our method obtains a large gain in robust acc. without compromising much on clean performance and takes significantly lesser computational overhead compared to robust few-shot methods.}
\label{fig:motivation}
\end{figure*}
This work handles the above concerns by proposing a simple but effective alternative that avoids complicated training like meta-learning and does not require any adversarial samples in its training. Our approach for robust few-shot learning is inspired by human cognition. Humans primarily rely on low frequencies of a given image to associate it with one of the target labels~\cite{wang2020high,wang2020towards}. However, 
deep models are tuned to obtain high accuracy, and in that process, they become sensitive to high-frequency regions of the image. Hence perturbations in those regions can easily impact the model’s decision while remaining human imperceptible. These high frequency regions gets heavily contaminated by adversarial attacks (refer Sec. IV-A in supplementary). Hence, to resist these models from being fooled by adversarial images and enforce similar behavior to humans in decision-making, we explicitly force the network to learn on low frequencies. However, high frequencies allow deep models to be more discriminative, necessary for good clean accuracy. So, ensuring the models trained on low frequencies as discriminative as those trained with original images can yield good accuracy on both clean and adversarial images. With this intuition, our proposed approach has broadly two stages: Pretraining and Finetuning with evaluation (as shown in Fig.~\ref{fig:proposed}).

In the pretraining stage, we first train a Teacher network on base classes, which is used for self-distillation. 
The Student network is then trained on low frequencies of original samples while matching discriminative features of the Teacher network on corresponding original data via cosine similarity. In the finetuning stage, similar to~\cite{Dhillon2020A} we transductively finetune the entire pretrained network. To boost the robustness, we additionally apply cosine similarity loss on the features of the unlabelled query set and its corresponding low frequencies. We observed that deep models find it hard to learn rich discriminative features when trained directly on very low-frequency images. We further empirically found that progressively moving from low-frequency samples obtained at high-radius to low-radius facilitates deep model training to be more discriminative while still being low-frequency biased. 
The low-frequency samples are computed via taking inverse Fast Fourier Transform of the frequency representation obtained by masking frequency components greater than selected radius (initially high and progressively made lower during training).

Finally, the finetuned model is evaluated using a weighted logits ensemble of network predictions on diverse low-frequency samples
. The utility of our proposed approach is shown in Fig.~\ref{fig:motivation}. We obtain significantly better adversarial accuracy than existing few-shot robust methods, without losing much on clean accuracy compared to non-robust methods. Also, our training time is much lower than existing meta-learning-based robust methods. For e.g., we beat the state-of-the-art robust few-shot method (AQ-R2D2) by $12.6\%$ in clean and $28.2\%$ in robust accuracy with $\approx$ $\times5$ faster training time i.e. only taking 5400.11 seconds compared to 27582.44 taken by AQ-R2D2. 

The overall contributions are summarized below:
\begin{itemize}
\item We are the first to propose a computationally cheaper non-meta learning approach for robust few-shot learning that does not require any adversarial samples.
\item We train networks to enforce similar decision-making as human cognitive processes to incorporate robustness. For this, we perform pretraining using self-distillation via cosine similarity to make the feature representation of low-frequency samples close to original samples of base classes. Similarly, we also apply this loss on the query set and their low-frequency counterparts while finetuning on novel classes that further boost the robustness. 
\item The model trained with low frequencies obtained via FFT at low or high radius results in poor performance in either clean or adversarial samples. To avoid that, we progressively apply cosine similarity loss on the low-frequency samples obtained by varying radius from an easy high radius (favoring more high frequencies) to gradual hard low radii (favoring less high frequencies). 
\item We demonstrate the efficacy of our proposed framework via extensive ablations and experiments on benchmark datasets for different few-shot settings (sec.~\ref{sec:experiments}).
\end{itemize}
\section{Related Works}
\textbf{Adversarial defense:} 
Adversarial Training (AT) has proved to be the most effective and reliable technique which generates adversarial samples during training using min-max optimization. Goodfellow~\textit{et al.}~\cite{goodfellow2014explaining} proposed the first AT method that crafted the adversarial examples by Fast Gradient Sign Method (FGSM) attack. Following this~Madry~\textit{et al.}~\cite{madry2017towards} reported improvement in robustness performance by using a stronger attack namely Projected Gradient Descent (PGD) in AT. Later on,~Zhang~\textit{et al.}~\cite{zhang2019theoretically} further regularized AT by using a KL divergence loss term to enforce similar output distribution on clean and adversarial data. These additional regularizations in AT methods can be viewed similar to self-distillation ($S\rightarrow S$ in Sec.~\ref{subsec:teacher_network}) but relies on adversarial samples. With explicit use of large Teacher ($T$) network, robustness is transferred to lightweight student ($S$) in ~\cite{goldblum2020adversarially,zi2021revisiting}. However, the network T is assumed to be adversarially trained and robust to adversarial attacks unlike ours where $T$ is a non-robust network trained with standard cross entropy and $S$ has identical architecture as $T$. The AT methods usually have high computational complexity and also require high-capacity networks~\cite{zi2021revisiting}. An alternative to these, some Non AT based methods like JARN~\cite{chan2019jacobian}, BPFC~\cite{Addepalli_2020_CVPR}, GCE~\cite{chen2019improving} etc., have been proposed. They do not rely on adversarial samples and hence take less training time. 
However, they are not robust to a wide range of attacks and their performance is lower when faced with stronger attacks \cite{heattack2017}.

\textbf{Few-Shot classification:} 
One of the earliest and simplest baselines for few-shot classification consists of pre-training the model on the base classes and then finetuning on the novel-classes support set~\cite{bengio2012deep,chen2019closerfewshot}. Dhillon~\textit{et al.}~\cite{Dhillon2020A} demonstrated that transductive finetuning outperforms various state-of-the-art techniques across various benchmark datasets. Recently, the few-shot-learning community has shifted its attention to meta-learning, which attempts to ‘learn’ the ‘learning-algorithm’ itself over a distribution of tasks, such that it can quickly adapt to novel tasks using finetuning~\cite{snell2017prototypical,finn2017model, bertinetto2018meta, lee2019meta}. 

\textbf{Robust Few-Shot learning:} Current state-of-the-art adversarial robustness techniques prove to be ineffective in a few-shot setting due to their high dependence on large-labeled datasets~\cite{yin2018adversarial,goldblum2020adversarially}. Yin~\textit{et al.}~\cite{yin2018adversarial} proposed a novel meta-learning algorithm that facilitates the learning of accurate and robust few-shot learners. Goldblum~\textit{et al.}~\cite{goldblum2020adversarially} proposed an algorithm-agnostic adversarial robustness technique (Adversarial Querying) that can be combined with the existing state-of-the-art meta-learning techniques. Liu~\textit{et al.}~\cite{liu2021longterm} proposed a robust meta-learning (LCAT) that halves the training time compared to current Adversarial Querying while maintaining similar performance. 

The aforementioned techniques provide robustness by generating adversarial samples at the training stage which leads to high-computational costs. Moreover, these methods are primarily designed for meta-learning based solutions that involve complicated optimization (e.g. difficult to optimize for larger architectures \cite{Dhillon2020A}). 
In contrast to existing works, our proposed method is a non-meta learning based approach which is cheaper and effective while 
not depending on any adversarial samples (at neither pretraining nor finetuning).

Recently, several works~\cite{wang2020high,wang2020towards} observed that deep neural networks rely heavily on the high-frequency spectrum for prediction, making them highly sensitive to any adversarial perturbations in those regions. 
Li~\textit{et al.}~\cite{Li2022robust} further observed that many publicly available robust models showed a preference for low-frequency spectrum. 
Inspired by the above findings and to mimic human cognition in deep neural networks, we 
employ 
frequency regularisation to enforce few shot learners to rely on the low-frequency spectrum while ensuring 
model features to highly discriminative, resulting in correct predictions.
\section{Preliminaries}
\label{sec:prelim}
\textbf{Notations:} In the pretraining stage, we use base class data $D_{B}=\{(x_{i}^{B}, y_{i}^{B})\}_{i=1}^{b}$ containing $b$ samples belonging to one of the $c_{1}$ classes. The novel class data $D_{N}=\{D_{N_{s}}, D_{N_{q}}\}$ whose samples are from $c_{2}$ different classes, is used in the finetuning stage. $D_{N_{s}} =\{(x_{i}^{N_{s}}, y_{i}^{N_{s}})\}_{i=1}^{n_{s}}$ is the support set  consisting of $n_{s}$ labelled samples. $D_{N_{q}}= \{x_{i}^{N_{q}}\}_{i=1}^{n_{q}}$ is the query set with $n_{q}$ unlabelled samples. 

The \textit{Teacher} and \textit{Student} models are represented by $T$ and $S$ whose output $T(x; \theta)$ and $S(x; \theta)$ are the logits for any input image $x$ that is fed to the networks $T$ and $S$ respectively. We denote the softmax function by $P_{soft}(.)$.

$A_{adv}$ denote a set of $m$ distinct adversarial attacks i.e. $A_{adv} = \{A_{k}\}_{k=1}^{m}$. An $i^{th}$ adversarial image of query set i.e. $\tilde{x}_{i}^{N_{q}}$, is obtained by using an adversarial attack $A_{k} \in A_{adv}$  such that it fools the model $S$. $FT(.)$ and $FT^{-1}(.)$ denote fourier and inverse fourier transforms. For any $i^{th}$ sample ($x_{i}$) in the spatial domain, its corresponding representation in the frequency domain is represented by $z_{i}$. The low and high frequencies of $z_{i}$ separated at a radius $r$ are denoted by $zl_{ir}$ and $zh_{ir}$ whose spatial representations are $xl_{ir}$ and $xh_{ir}$ respectively.

\textbf{Adversarial attacks}: An attack $A_{k} \in A_{adv}$, carefully crafts noise ($\delta$) such that this noise when added to the input image ($x$) is human-imperceptible but the network predictions on them change. In other words, a model $M$ on input ($\tilde{x}$) gets fooled when $\mathrm{argmax}(M(x)) \neq \mathrm{argmax}(M(\tilde{x}))$ where $\tilde{x} = x + \delta$ and $\norm{\delta} \leq \epsilon$. We consider the popular $l_{\infty}$ threat model in our setup i.e. the  $l_{\infty}$ norm of the perturbations ($\norm{\delta}_{\infty}$) is within the $\epsilon$ ball.

\textbf{Fourier Transform}: This transformation establishes a mapping between spatial and frequency domain. To obtain frequency representation of an $i^{th}$ sample from its spatial domain, $FT(.)$ is applied i.e. $z_{i} = FT(x_{i})$. The frequencies in $z_{i}$ can be further split into low and high at a radius $r$ (note that the different strategies for selecting the value of radius $r$ is also discussed in Sec.~\ref{subsection: PL}) using Hadamard product with corresponding low and high frequency masks as shown below:
\begin{equation}
\begin{aligned}
zl_{ir} = z_{i} \circ F^{mask_{lr}} \hspace{0.2in} zh_{ir} = z_{i} \circ F^{mask_{hr}}
\end{aligned}
\end{equation}
The value of $F^{mask_{lr}}$ at a $p^{th}$ row and $q^{th}$ column is calculated as:  
\begin{equation}
\begin{aligned}
F_{pq}^{mask_{lr}} = \mathds{1}(\sqrt{(p-d/2)^{2} + (q-d/2)^{2}}) < r)
\end{aligned}
\end{equation}
where $\mathds{1}$ is an Indicator function. 
The zero frequency of the sample ($z_{i}$) obtained via $FT(.)$ is shifted to the center ($d/2$,$d/2$) where $d \times d$ is the dimension of $z_{i}$. The value of low-frequency mask at a location ($p$,$q$) is $1$ if its euclidean distance from the center is less than $r$, otherwise the value is $0$. $F^{mask_{hr}}$ is the complement of $F^{mask_{lr}}$. To get the spatial representation of the above frequencies:
\begin{equation}
\begin{aligned}
xl_{ir} = FT^{-1}(zl_{ir}) \hspace{0.2in} xh_{ir} = FT^{-1}(zh_{ir})
\end{aligned}
\end{equation}

\textbf{Self Distillation}: Transfer of knowledge from network T to network S having identical architectures.

\textbf{Few-Shot setting}: During the training stage, the model is trained from scratch on $D_{B}$ data containing base classes. During testing, the pretrained model is finetuned on a small sized support set $D_{N_{s}}$ ($n_{s} \ll b$) that has samples from novel classes. For a $k$-way $n$-shot set up, the following holds: $c_{2} = k$,  $n_{s} = k \cdot n$. 
We follow the setup described by~\cite{goldblum2020adversarially} where the finetuned model is susceptible to adversarial attacks and the query set is perturbed with an objective to fool the model. 
The robustness of the finetuned model is then quantified by evaluating it on both the clean query set 
$\{x^{N_{q}}\}_{i=1}^{n_{q}}$ and the query perturbed samples $\{\tilde{x}^{N_{q}}\}_{i=1}^{n_{q}}$.
\begin{figure*}[htp]
\centering
\centerline{\includegraphics[width=\textwidth]{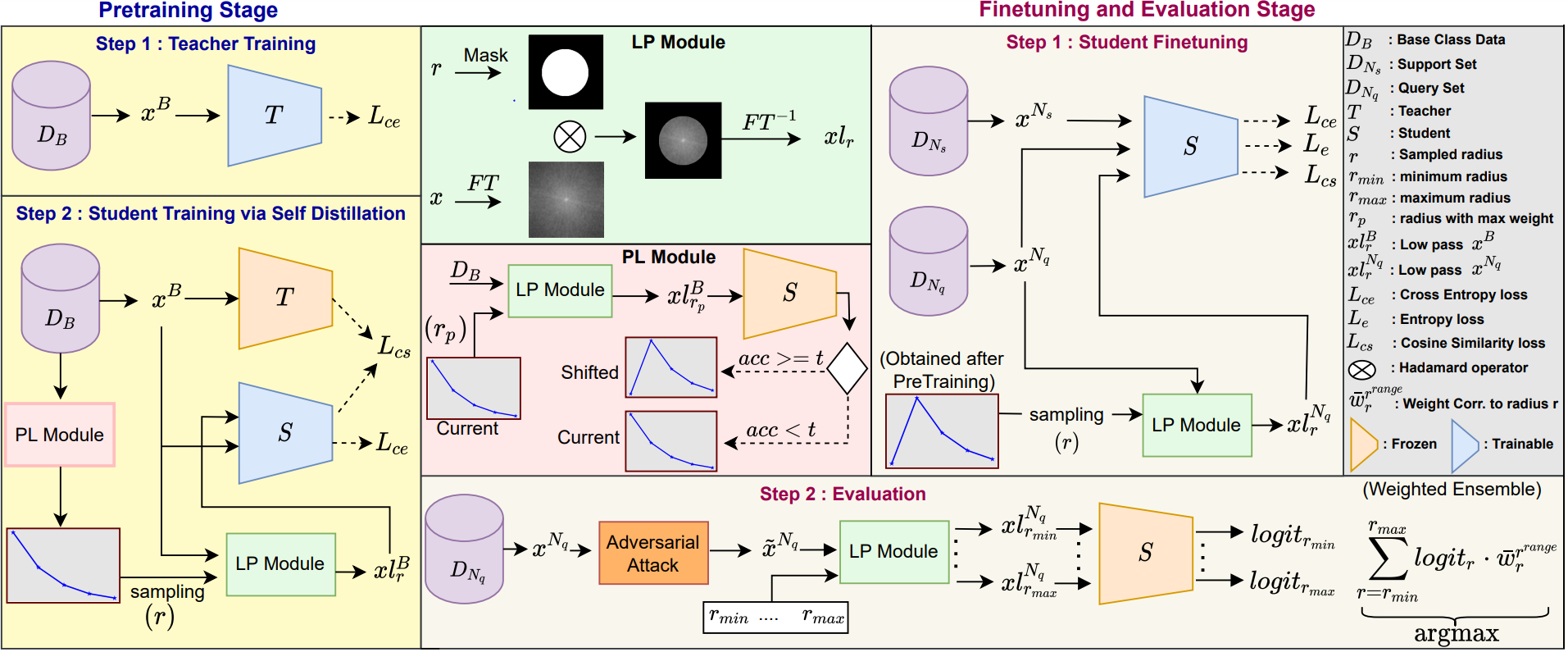}}
\caption{
Detailed steps of our proposed approach for robust few-shot learning. In pretraining stage, the teacher network ($T$) is trained on base class data $D_{B}$ (step $1$). The student network ($S$) is trained (step $2$) via self distillation using the trained network $T$. The PL module uses an initial long tail weight distribution to sample radius. The distribution is shifted if student performance on the peak radius ($r_{p}$) crosses the threshold ($t$). In the LP module, the low-pass filtered version of an input image 
is computed using a low-frequency mask (at sampled radius `$r$') in the Fourier transform ($FT$), followed by $FT^{-1}$. In finetuning stage (step $1$), apart from the $L_{ce}$ on support set ($x^{N_{s}}$) and $L_{e}$ loss on query set ($x^{N_{q}}$), we additionally apply $L_{cs}$ loss on ($x^{N_{q}}$). The weighted distribution ($\bar{w}^{r^{range}}$) over radii via PL module obtained at the end of pretraining is used in this stage to sample radii at which low frequencies are obtained via LP module. Finally, in the evaluation step
, the logits obtained at each radii ($r_{max}$ to $r_{min}$) are weighted using the weight distribution $\bar{w}^{r^{range}}$. The complete training in both the stages does not require generation of any adversarial sample, 
making the method computationally efficient.}
\label{fig:proposed}
\end{figure*}
\section{Proposed Approach}
\label{sec:proposed}
Our proposed approach is broadly divided into two stages: a)Pretraining and b)Finetuning and Evaluation. In pretraining stage, we first train the teacher network on $D_{B}$ and then use it for self distillation. The student network is trained on low-frequency samples of $D_{B}$ to match the feature responses of the teacher model on original samples. In second stage, the pretrained model is finetuned transductively on $D_{N}$ along with our additional loss to improve dicriminativeness on low-frequency query samples. The finetuned model is then evaluated on both the clean and adversarially pertrubed query samples using the logits obtained at different low-frequencies of the input. The proposed method is also described in Fig.~\ref{fig:proposed}. Next, we discuss each of these stages in detail.

\textbf{Stage1: Pretraining}: In the first step, we train the network $T$ from scratch on $D_{B}$ by minimizing the cross entropy loss ($L_{ce}$) as shown below:
\begin{equation}
\begin{aligned}
\theta_{t}^{*} = \min_{\theta_{t}} L_{ce} = \min_{\theta_{t}} \frac{1}{b} \sum_{i=1}^{b} -log (P_{soft}(T(x_{i}^{B}))_{y_{i}^{B}})
\end{aligned}
\end{equation}
$L_{ce}$ loss ensures that network $T$ learns highly discriminative features on the base class data. In the second step, we use the trained network $T$ from step $1$ as a Teacher model to train the network $S$ via self distillation. Networks $T$ and $S$ are of identical architectures where $S$ is initialized with $\theta_{t}^{*}$. The network $T$ is used to obtain logit responses on samples of $D_{B}$. The network $S$ is trained on low frequencies of $D_{B}$ samples to match the logit responses via maximization of cosine similarity loss ($L_{cs}$) as mentioned below along with $L_{ce}$ loss on $D_{B}$:
\begin{equation}\label{eq5}
\begin{gathered}
\theta_{s}^{*}=\min_{\theta_{s}}(L_{ce}-L_{cs})\text{;}\hspace{0.05in}
L_{cs}=\frac{1}{b} \sum_{i=1}^{b}  \frac{S(xl_{i}^{B})^TT(x_{i}^{B})}{ \norm{S(xl_{i}^{B})} \norm{T(x_{i}^{B})} }
\end{gathered}
\end{equation}
Distinguishing class labels using low-frequency samples is a significantly challenging task compared to vanilla classification as low-frequency samples inherently contain less information. Hence, directly training the student network $S$ on low-frequency images using $L_{ce}$ would lead to poor performance. We overcome this challenge by using the $ L_{cs}$ loss which encourages network $S$ to be as discriminative on low-frequency samples as the network $T$ is for original samples in the feature (logits) space. To obtain $xl^{B}$ in order to compute this loss above, we first convert the spatial data into frequency domain using $FT$ i.e. $z_{i}^{B} = FT(x_{i}^{B}), \forall i \in [1..b]$. Next, low-frequency samples ($xl^{B}$) are obtained by superimposing a low frequency mask (at a radius $r$) on the frequency domain representation followed by $FT^{-1}$. Refer Fourier Transform operations in Sec.~\ref{sec:prelim}.

Generating the low-frequency mask at a suitable radius $r^{*}$ is crucial for achieving good robustness as, if $r^{*}$ is too low it could lead to lower performance due to less discriminability (even if no adversarial contamination is there) and if $r^{*}$ is too high it could lead to lower performance due to adversarial contamination (even if more discriminability is there. However, empirically estimating $r^{*}$ is impractical without generating adversarial samples as it highly depends on the dataset, model, etc. In such a scenario, an intuitive approach could be to select the lowest radius possible (to ensure less adversarial contamination) while maintaining  discriminativeness (to ensure decent clean accuracy). For that, we perform a weighted sampling over the radii defined on range ([$r_{max}$, $r_{min}$]) where weights form a long-tailed distribution (described below) by initially assigning  a higher weight to $r_{max}$ as it’s relatively easier to learn at high radius.
\begin{equation}\label{eq6}
\begin{split}
w^{r^{range}}_{r} = \lambda^{i}  \hspace{0.05in}, \hspace{0.05in} \text{such that} \\\hspace{0.05in} \lambda \in (0,1) \text{,}  \hspace{0.1in} r= (r^{range})_{i} \hspace{0.05in} \text{and} \hspace{0.05in} \\r^{range} = [r_{max}, r_{max}-1,..., r_{min}]
\end{split}
\end{equation}
Here, $i$ is the index of the radius $r \in r^{range}$. The stochastic weighted sampling also encourages the model to avoid overfitting on samples of any particular radius. Once training accuracy of the model on low-frequency samples (at $r_{max}$) exceeds the threshold ($t$), we shift the weighting scheme by having the highest weight at radius $r_{max}-1$. We repeat this process till $r_{p}$ (radius corresponding to the highest weight) is equal to $r_{min}$. We term this technique as ‘Progressive Learning’ (PL), as we learn discriminative features on low-frequency samples by gradually moving from high to low radius.

\textbf{Stage 2: Finetuning and Evaluation}: A new classification layer having cosine softmax with $k$ class nodes for a $k$-way $n$-shot setting is added on top of the logits of pretrained model $S$ obtained from the previous stage. The weights of this layer are initialized with class mean features of the support set where the mean feature vector corresponding to each class is taken as a row in the weight matrix with biases as zeros. This layer’s inputs and weights are normalized before taking softmax on their dot product. We perform finetuning on this model using samples of $D_{N}$ with losses in eq.~\ref{eq7}.
\begin{equation}\label{eq7}
\begin{gathered}
L_{ce}= 1/n_{s}  \sum_{i=1}^{n_{s}} -log P_{soft}(S(x_{i}^{N_{s}}))_{y_{i}^{N_{s}}} \\ \noalign{\vspace{-0.03in}}
L_{e} = 1/n_{q}  \sum_{i=1}^{n_{q}}   \sum_{j=1}^{k} -P_{soft}(S(x_{i}^{N_{q}}))_{j} \hspace{0.05in}log P_{soft}(S(x_{i}^{N_{q}}))_{j} \\ \noalign{\vspace{-0.03in}}
L_{cs} = 1/n_{q}  \sum_{i=1}^{n_{q}} \frac{S(xl_{ir}^{N_{q}})^T S(x_{i}^{N_{q}})}{\norm{S(xl_{ir}^{N_{q}})} \norm{S(x_{i}^{N_{q}})} }
\end{gathered}
\end{equation}
The discriminative low-frequency features learned during pre-training on $D_{B}$ might not be directly useful during finetuning on $D_{N}$ as $D_{B}$ and $D_{N}$ consist of disjoint sets of classes. Thus, we apply $L_{cs}$ loss on the query-set samples in the finetuning stage to improve discriminativeness of low-frequency features of novel class samples. 
Further, evaluating the performance on low-frequency samples of a small-sized support set is not a good indicator for updating the weight distribution. Even if the query data is relatively more, we cannot use it to evaluate the performance to compare with threshold, as it is unlabeled. Hence, rather than following a similar progressive learning strategy as pretraining, we employ the weight distribution ($\bar{w}^{r^{range}}$) obtained at the end of pretraining to generate the corresponding low-frequency samples. 
The finetuned model $S$ with optimal parameters (i.e. $\theta_{s}^{'} = \min_{\theta_{s}} (L_{ce} + L_{e} - L_{cs})$), is used to predict labels on the query set $D_{N}^{q}$. During evaluation on a query sample, we perform a weighted-average of the logits predicted on low-frequency input at each radius $r \in [r_{min}, r_{max}]$.
\begin{equation}
    Pred = \mathrm{argmax}(\sum_{r=r_{min}}^{r_{max}} (logit_{r} \cdot \bar{w}_{r}^{r^{range}}))
\end{equation}
Here, $\bar{w}_{r}$ is the weight corresponding to 
radius $r$ in $r^{range}$ using the weight distribution obtained at the end of pretraining and $logit_{r}$ is the logits when low frequency sample ($xl_{r}^{N_{q}}$) obtained at the 
radius $r$ is fed to the finetuned model $S$.
All the steps related to different stages (Pretraining and Finetuning with evaluation) that are discussed here, are also structurally summarized to provide an overall algorithm which is presented in supplementary material (Sec. V). Also, we qualitatively demonstrate the working of our method on clean and adversarially perturbed input in Sec. IV-B of supplementary. Next we validate our proposed method through extensive experiments via different ablations and comparison with state-of-the-art. 
\section{Experiments}
\label{sec:experiments}
We demonstrate the effectiveness of our technique by performing experiments on two benchmarks datasets in few-shot-learning, specifically, CIFAR-FS~\cite{bertinetto2018meta} and Mini-ImageNet~\cite{vinyals2016matching}. As per the standard protocol, we use $15$ as query shot and report the estimate of the mean accuracy evaluated over $1000$ few-shot episodes along with the $95\%$ confidence interval of this estimate. 
We also evaluate our method in more challenging realistic scenarios, such as considering an unequal amount of samples across classes in a support set, or a varying query set size. In such cases, we also observe similar consistent performance 
We use the ResNet-12~\cite{he2016deep} architecture as our backbone feature extractor for all the experiments and ablations. Refer to supplementary (Sec. I-B) where we also perform an ablation over different choices of backbone architectures and observe that our method consistently improves the adversarial accuracy across architectures compared to the respective baselines. Similar to \cite{Dhillon2020A}, our baseline (vanilla) model consists of pretraining using $L_{ce}$ and mixup-regularization, and finetuning using $L_{ce}$ (on support set) and $L_{e}$ (on query set). 
 The value of radius is kept fixed at $2$ i.e. $r=2$  for ablations pertaining to $L_{cs}$ and the evaluation is performed on the same $r$ unless otherwise specified. In the subsequent subsections, we first validate the utility of the various components used in Pretraining and Finetuning stages of our method (Sec.~\ref{subsec:frequency_reg} to Sec.~\ref{subsection: PL}), followed by comparison with state-of-the-art methods (Sec.~\ref{subsection: SOTA})). For additional training details refer supplementary (Sec. III)
\subsection{Frequency Regularization in Pretraining}
\label{subsec:frequency_reg}
\vspace{-0.05in}
\begin{table}[htp]
\caption{Benefits of adding frequency regularisation (using $L_{cs}$ loss) in pretraining on CIFAR-FS for Resnet$12$. The robustness against both PGD and Auto Attack is significantly improved but at the cost of clean accuracy.
}
\centering
\scalebox{0.95}{
\begingroup
\setlength{\tabcolsep}{0.15cm} 
\begin{tabular}{|c|c|c|c|c|}
\hline
Setup            & Method     & Clean & PGD         & Auto Attack \\ \hline 
\multirow{2}{*}{1 - SHOT} & Vanilla ($L_{ce}$)  & 68.40 (0.71)   & 0.01 (0.01)          & 0.00 (0.00)          \\ \cline{2-5} 
                          & \cellcolor{gray!20}{$L_{ce}$ - $L_{cs}$} & \cellcolor{gray!20}{22.42 (0.26)}   
                          &\cellcolor{gray!20}{\textbf{22.40 (0.26)}} & \cellcolor{gray!20}{\textbf{22.32 (0.26)}} \\ \Xhline{2\arrayrulewidth}
\multirow{2}{*}{5 - SHOT} & Vanilla ($L_{ce}$)  & 81.81 (0.52)   & 0.07 (0.02)          & 0.00 (0.00)          \\ \cline{2-5} 
                          & \cellcolor{gray!20}{$L_{ce}$ - $L_{cs}$} & \cellcolor{gray!20}{25.09 (0.36)}   & \cellcolor{gray!20}{\textbf{24.11 (0.36)}} & \cellcolor{gray!20}{\textbf{25.01 (0.36)}} \\ \hline
\end{tabular}
\endgroup
}
\label{tab:pretrain-low-freq}
\end{table}
We motivate the benefit of regularizing the vanilla network ($S$) to match the feature responses (logits) of the original and low-frequency samples during pre-training in Table~\ref{tab:pretrain-low-freq}. The vanilla network achieves decent $1$-SHOT and $5$-SHOT clean accuracies on the CIFAR-FS dataset, however, it’s extremely vulnerable to adversarial perturbations. Thus, we regularize $S$ by matching the logits of an original and its corresponding low-frequency sample using $L_{cs}$. The aim of such frequency-based regularization is to reduce the dependency of the model on high-frequency components. We observe that such frequency-based regularization indeed leads to a non-trivial increase in robust accuracy (across different attacks). However, there is a huge drop in the clean accuracy as low-frequency samples inherently have less discriminative content (at the cost of reducing adversarial perturbations/contamination) compared to original samples.
\subsection{Utility of Teacher Network}
\label{subsec:teacher_network}
\begin{table}[htp]
\caption{Comparison of student ($S$) performance when distilled from itself (S$\rightarrow$S) and distilled from teacher (T$\rightarrow$S). The T$\rightarrow$S distillation gives almost $2\times$ better mean accuracy on both clean and adversarial samples.
}
\centering
\scalebox{0.93}{
\begingroup
\setlength{\tabcolsep}{0.15cm} 

\begin{tabular}{|c|c|c|c|c|}
\hline
Setup                     & Method                                                                 & Clean        & PGD          & Auto Attack  \\ \hline
\multirow{2}{*}{1 - SHOT} & $L_{ce} $ - $L_{cs}$ (S$\rightarrow$S) & 22.42 (0.26) & 22.40 (0.26) & 22.32 (0.26) \\ \cline{2-5} 
                          & \cellcolor{gray!20} $L_{ce} $ - $L_{cs}$ (T$\rightarrow$S) & \cellcolor{gray!20} \textbf{45.01 (0.63)}  & \cellcolor{gray!20} \textbf{44.16 (0.63)} & \cellcolor{gray!20} \textbf{44.28 (0.63)}  \\ \Xhline{2\arrayrulewidth}
\multirow{2}{*}{5 - SHOT} & $L_{ce} $ - $L_{cs}$ (S$\rightarrow$S) & 25.09 (0.36) & 24.11 (0.36) & 25.01 (0.36) \\ \cline{2-5} 
                          &\cellcolor{gray!20} $L_{ce} $ - $L_{cs}$ (T$\rightarrow$S) &\cellcolor{gray!20} \textbf{53.86 (0.58)} & \cellcolor{gray!20} \textbf{52.71 (0.57)}  &\cellcolor{gray!20} \textbf{52.90 (0.57)} \\ \hline
\end{tabular}
\endgroup
}
\label{tab:teacher}
\end{table}
In an attempt to increase the model’s discriminability on low-frequency samples, we draw inspiration from the knowledge-distillation literature. Specifically, we first train a teacher model ($T$) using $L_{ce}$ and employ it in a self-distillation setup in order to increase the student model’s ($S$) discriminability. $S$ is trained using a combination of $L_{ce}$  and $L_{cs}$ loss as in eq.~\ref{eq5}. We further fix the batchnorm layer parameters of S after initializing S with T weights. The rich fine-grained knowledge from $T$ assist $S$ to learn more discriminative features even for low-frequency samples. In Table~\ref{tab:teacher} we observe a significant boost in performance (across all the metrics) for the $T$$\rightarrow$$S$ self-distillation setup compared to the previous setting where the low-frequency and original samples logits are obtained from the same $S$ network. Although, at par with most state-of-the-art few-shot robustness techniques, there is still a significant drop in clean accuracy (compared to the vanilla network) for the $T$$\rightarrow$$S$ setup. 

Once, the $S$ model is trained on $D_{B}$ using the $T$$\rightarrow$$S$ self-distillation setup as described in stage $1$ we shift our focus to improvements during finetuning stage to further boost clean and robust accuracies.
\subsection{Frequency Regularization in Finetuning}
\vspace{-0.05in}
\begin{table}[htp]
\caption{Frequency regularization during finetuning yields significant improved performance (mean acc.).
}
\centering
\scalebox{0.95}{
\begingroup
\setlength{\tabcolsep}{0.15cm} 

\begin{tabular}{|c|c|c|c|c|}
\hline
Setup                     & Method (Stage 2)                & Clean        & PGD          & Auto Attack  \\ \hline
\multirow{2}{*}{1 - SHOT} & $L_{ce}$ + $L_{e}$             & 45.01 (0.63) & 44.16 (0.63) & 44.28 (0.63) \\ \cline{2-5} 
                          & \cellcolor{gray!20} $L_{ce}$ + $L_{e}$ - $ L_{cs}$ & \cellcolor{gray!20} \textbf{61.49 (0.70)} & \cellcolor{gray!20} \textbf{60.73 (0.70)}  & \cellcolor{gray!20} \textbf{60.83 (0.70)}  \\ \Xhline{2\arrayrulewidth}
\multirow{2}{*}{5 - SHOT} & $L_{ce}$ + $L_{e}$             & 53.86 (0.58) & 52.71 (0.57) & 52.90 (0.57) \\ \cline{2-5} 
                          & \cellcolor{gray!20}$L_{ce}$ + $L_{e}$ - $ L_{cs}$ & \cellcolor{gray!20} \textbf{73.17 (0.60)} & \cellcolor{gray!20} \textbf{72.26 (0.61)} & \cellcolor{gray!20} \textbf{72.53 (0.60)} \\ \hline
\end{tabular}
\endgroup
}
\label{tab:L_cs_finetuning}
\end{table}
Motivated by the effectiveness of $L_{cs}$ during pre-training we also apply $L_{cs}$ during finetuning. Since, $L_{cs}$ can be applied in an unsupervised manner (no label information required) we directly optimize it on the query-set, in addition to $L_{ce}$ and $L_{e}$ losses (as shown in eq.~\ref{eq7}). In Table~\ref{tab:L_cs_finetuning}, we observe that adding $L_{cs}$ at finetuning stage leads to impressive gains as both the clean and robust accuracy increase by $\approx 15\%$ and $20\%$ in $1$-SHOT and $5$-SHOT settings respectively. We also performed additional 
ablations in supplementary such as different options for applying $L_{cs}$ loss (Sec. I-C) and choices of loss functions for frequency regularization (Sec. I-A).
\subsection{Effectiveness of Progressive Learning}
\label{subsection: PL}
\begin{table}[htp]
\caption{Performance (mean accuracy in \% with confidence interval) comparison of our proposed Progressive Learning (PL) strategy with different radius selection techniques. Our PL technique based on dynamic weighted sampling not only overcomes the sensitivity in the model performance associated with choice of a fixed radius but also leads to significantly better adversarial performance than uniform sampling over radii. 
}

\centering
\scalebox{0.95}{
\begingroup
\setlength{\tabcolsep}{0.15cm} 

\begin{tabular}{|c|c|c|c|c|}
\hline
Setup                   & Method                                             & Clean        & PGD                   & Auto Attack  \\ \hline
\multirow{4}{*}{1-SHOT} & Fixed ($r$=2)                                      & 61.49 (0.70) & \textbf{60.73 (0.70)} & 60.83 (0.70) \\ \cline{2-5} 
                        & Fixed ($r$=16)                                     & 63.48 (0.69) & 1.99 (0.19)           & 3.19 (0.18)  \\ \cline{2-5} 
                        & r$\sim$ $\mathcal{U}\{2,16\}$ & 63.90 (0.70) & 52.63 (0.80)          & 55.80 (0.77) \\ \cline{2-5} 
 & \begin{tabular}[c]{@{}c@{}} \cellcolor{gray!20} PL on $r$ [16,..,2] \\ \cellcolor{gray!20} (\textbf{Ours})\end{tabular} & \cellcolor{gray!20} \textbf{65.03 (0.72)} & \cellcolor{gray!20} 60.55 (0.76)          & \cellcolor{gray!20} \textbf{62.05 (0.74)} \\ \Xhline{2\arrayrulewidth}
\multirow{4}{*}{5-SHOT} & Fixed ($r$=2)                                      & 73.17 (0.60) & 72.26 (0.61)          & 72.53 (0.60) \\ \cline{2-5} 
                        & Fixed ($r$=16)                                     & 78.81 (0.53) & 3.46 (0.17)           & 5.24 (0.23)  \\ \cline{2-5} 
                        & r$\sim$ $\mathcal{U}\{2 ,16\}$ & 79.06 (0.54) & 68.71 (0.71)          & 71.69 (0.67) \\ \cline{2-5} 
 & \begin{tabular}[c]{@{}c@{}} \cellcolor{gray!20}PL on $r$ [16,..,2] \\ \cellcolor{gray!20} (\textbf{Ours})\end{tabular} & \cellcolor{gray!20} \textbf{79.79 (0.55)} & \cellcolor{gray!20} \textbf{75.93 (0.61)} & \cellcolor{gray!20} \textbf{77.08 (0.59)}

 \\ \hline
\end{tabular}
\endgroup
}
\label{tab:ablation-radius}
\end{table}
The previously described ablations were performed at a fixed radius $r=2$. Now, we explain the efficacy of our progressive learning technique where we set $r_{min}$ to $2$ and $r_{max}$ to $N/2$ where $N$ is the largest frequency component for a given dataset. As discussed in Sec.~\ref{sec:proposed} both clean and adversarial performance significantly suffer if $r^{*}$ is either too high (implies high adversarial contamination) or too low (implies low discriminativeness). We empirically validate this phenomenon by repeating our proposed frequency regularization method at two distinct radii i.e. $r=2$ (low) and $r=16$ (high) (shown in Table~\ref{tab:ablation-radius}) and observe that although there is an increase in clean accuracy for $r=16$ as compared to $r=2$, there is a huge drop in robust accuracy. This sensitivity on the radius $r$ motivates us to propose a data-driven method of selecting/sampling $r$ during both pre-training and finetuning. We 
use our PL strategy to perform a dynamic weighted sampling over a broad range of radii [$r_{max}$, $r_{min}$] (as per eq.~\ref{eq6}). 
We observe that PL-based frequency regularization leads to a further boost in clean and robust accuracy. Most importantly, it automates the process of selecting the radius in a data-driven way. We also experiment with a baseline ensembling technique that uniformly samples radii between [$2$, $N/2$] and performs ensembling (equal weight to each radius) on the same range. We note that the uniform sampling based setting significantly underperforms PL-based sampling across both $1$-SHOT/$5$-SHOT settings, especially on robust accuracies which highlights the efficacy of our PL technique. In addition to these, we also perform sensitivity analysis 
where we vary the values of $r^{range}$ (Sec. II in supplementary) and hyperparameter $\lambda$ used in weighted sampling (eq.~\ref{eq6}) (Sec.~\ref{subsection:lambda_sens}).

\begin{table*}[htp]
\caption{Comparison of proposed approach (Ours) with state-of-the-art robust few-shot methods 
on CIFAR-FS and Mini-ImageNet datasets, against adversarial attacks used in existing works. AQ \cite{goldblum2020adversarially} and LCAT \cite{liu2021longterm} results are reported from their respective papers. 
Our proposed method achieves significant robustness compared to baseline and comfortably outperforms the existing few-shot robust methods by a large margin on both clean and adversarial data. 
}
\centering
\scalebox{0.9}{
\begingroup
\setlength{\tabcolsep}{0.18cm} 
\renewcommand{\arraystretch}{1.3}
\begin{tabular}{|l|c|cccc||cccc|}
\hline
\multirow{3}{*}{\textbf{Method}} &
  \multirow{3}{*}{\textbf{\begin{tabular}[c]{@{}c@{}}No\\Adv.\\ Samples\end{tabular}}} &
\multicolumn{4}{c|}{\textbf{CIFAR-FS}} & \multicolumn{4}{c|}{\textbf{Mini-ImageNet}} \\ \cline{3-10} 

  & & \multicolumn{2}{c|}{\textbf{1 - SHOT}} &
  \multicolumn{2}{c|}{\textbf{5 - SHOT}}  & \multicolumn{2}{c|}{\textbf{1 - SHOT}} &
  \multicolumn{2}{c|}{\textbf{5 - SHOT}} \\ \cline{3-10} 
 &
   &
  \multicolumn{1}{c|}{Clean} &
  \multicolumn{1}{c|}{PGD} &
  \multicolumn{1}{c|}{Clean} &
  \multicolumn{1}{c|}{PGD} & \multicolumn{1}{c|}{Clean} &
  \multicolumn{1}{c|}{PGD} &
  \multicolumn{1}{c|}{Clean} &
  \multicolumn{1}{c|}{PGD} \\ \hline \hline
AQ - PROTO NET & \xmark
   & 
  \multicolumn{1}{c|}{42.33} &
  \multicolumn{1}{c|}{26.48} &
  \multicolumn{1}{c|}{63.53} &
  \multicolumn{1}{c|}{40.11} & 
  \multicolumn{1}{c|}{33.31} &
  \multicolumn{1}{c|}{17.69} &
  \multicolumn{1}{c|}{52.04} &
  27.99  \\ \hline
AQ - R2D2 & \xmark
   &
  \multicolumn{1}{c|}{52.38} &
  \multicolumn{1}{c|}{32.33} &
  \multicolumn{1}{c|}{69.25} &
  \multicolumn{1}{c|}{44.80} &
  \multicolumn{1}{c|}{37.91} &
  \multicolumn{1}{c|}{20.59} &
  \multicolumn{1}{c|}{57.87} &
  31.52 \\ \hline
AQ - MetaOptNet & \xmark
   &
  \multicolumn{1}{c|}{53.27} &
  \multicolumn{1}{c|}{30.74} &
  \multicolumn{1}{c|}{71.07} &
  \multicolumn{1}{c|}{43.79} &
  \multicolumn{1}{c|}{43.74} &
  \multicolumn{1}{c|}{18.37} &
  \multicolumn{1}{c|}{60.71} &
  28.08  \\ \hline
LCAT - ProtoNet & \xmark
   &
  \multicolumn{1}{c|}{42.51 (0.65)} &
  \multicolumn{1}{c|}{24.95 (0.61)} &
  \multicolumn{1}{c|}{61.69 (0.60)} &
  \multicolumn{1}{c|}{38.51(0.66)} &
  \multicolumn{1}{c|}{35.29 (0.51)} &
  \multicolumn{1}{c|}{19.65 (0.41)} &
  \multicolumn{1}{c|}{48.93 (0.53)} &
  29.08 (0.49) \\ \hline
LCAT - R2D2 & \xmark
   &
  \multicolumn{1}{c|}{41.49 (0.62)} &
  \multicolumn{1}{c|}{28.27 (0.59)} &
  \multicolumn{1}{c|}{54.97 (0.60)} &
  \multicolumn{1}{c|}{39.28 (0.61)} &
  \multicolumn{1}{c|}{26.47 (0.38)} &
  \multicolumn{1}{c|}{18.67 (0.33)} &
  \multicolumn{1}{c|}{39.77 (0.49)} &
  25.06 (0.43) \\ \hline
LCAT - MetaOptNet & \xmark
   &
  \multicolumn{1}{c|}{50.29 (0.70)} &
  \multicolumn{1}{c|}{24.13 (0.63)} &
  \multicolumn{1}{c|}{66.95 (0.58)} &
  \multicolumn{1}{c|}{35.22 (0.66)} &
  \multicolumn{1}{c|}{39.71 (0.54)} &
  \multicolumn{1}{c|}{12.51 (0.34)} &
  \multicolumn{1}{c|}{54.91 (0.50)} &
  18.92 (0.41)  \\ \hline
LCAT - Trades - ProtoNet & \xmark
   &
  \multicolumn{1}{c|}{37.94 (0.63)} &
  \multicolumn{1}{c|}{25.75 (0.58)} &
  \multicolumn{1}{c|}{51.14 (0.62)} &
  \multicolumn{1}{c|}{36.50 (0.62)} &
  \multicolumn{1}{c|}{29.58 (0.43)} &
  \multicolumn{1}{c|}{18.79 (0.37)} &
  \multicolumn{1}{c|}{39.53 (0.48)} &
  25.38 (0.43)  \\ \hline
LCAT - Trades - R2D2 & \xmark
   &
  \multicolumn{1}{c|}{35.88 (0.61)} &
  \multicolumn{1}{c|}{27.85 (0.56)} &
  \multicolumn{1}{c|}{47.14 (0.59)} &
  \multicolumn{1}{c|}{37.37 (0.59)} &
  \multicolumn{1}{c|}{25.88 (0.39)} &
  \multicolumn{1}{c|}{18.39 (0.35)} &
  \multicolumn{1}{c|}{32.28 (0.42)} &
  22.35 (0.38) \\ \hline
LCAT - Trades - MetaOptNet & \xmark
   &
  \multicolumn{1}{c|}{41.15 (0.65)} &
  \multicolumn{1}{c|}{29.51 (0.62)} &
  \multicolumn{1}{c|}{55.72 (0.61)} &
  \multicolumn{1}{c|}{41.33 (0.64)} &
  \multicolumn{1}{c|}{31.24 (0.46)} &
  \multicolumn{1}{c|}{18.51 (0.39)} &
  \multicolumn{1}{c|}{42.49 (0.51)} &
  26.18 (0.46) \\ \hline
   
  \hline \hline
    \rowcolor{Gray}
Baseline & \cmark
  &
  \multicolumn{1}{c|}{68.40 (0.71)} &
  \multicolumn{1}{c|}{0.01 (0.01)} &
  \multicolumn{1}{c|}{81.81 (0.52)} &
  \multicolumn{1}{c|}{0.07 (0.02)} &
  \multicolumn{1}{c|}{59.87 (0.67)} &
  \multicolumn{1}{c|}{0.00 (0.00)} &
  \multicolumn{1}{c|}{75.88 (0.51)} &
  0.00 (0.00) \\ \hline
     \rowcolor{Gray}
\textbf{Ours} & \cmark
  &
  \multicolumn{1}{c|}{65.03 (0.72)} &
  \multicolumn{1}{c|}{\textbf{60.55 (0.76)}} &
  \multicolumn{1}{c|}{79.79 (0.55)} &
  \multicolumn{1}{c|}{\textbf{75.93 (0.61)}} &
  \multicolumn{1}{c|}{54.05 (0.60)} &
  \multicolumn{1}{c|}{\textbf{53.63 (0.60)}} &
  \multicolumn{1}{c|}{66.33 (0.60)} &
  \textbf{65.80 (0.60)} \\ \hline
\end{tabular}
\endgroup
}
\label{tab:my-table}
\end{table*}

\subsection{Comparison with state-of-the-art methods}
\label{subsection: SOTA}
We compare the efficacy of our complete proposed method against various few-shot adversarial robustness techniques, namely, Adversarial Querying (AQ)
, Long-term Cross Adversarial Training 
 (LCAT)
 and LCAT+TRADES. 
Dhillon~\textit{et al.}~\cite{Dhillon2020A} use transductive finetuning where a combination of $L_{e}$ loss (on query set) and traditional $L_{ce}$ loss (on support set) is applied during finetuning. To have a fair comparison with them we 
follow the same protocol. However, they don’t evaluate performance on query perturbed samples, hence our setup is far more challenging than theirs (baseline). 
The results are reported in Table~\ref{tab:my-table}. Our method obtains significant gains in adversarial accuracy while retaining most of the clean accuracy compared to 
the baseline. On CIFAR-FS dataset, our method yields significant improvement of $\approx 11-29\%$ on clean data and $\approx 28-35\%$ on adversarial data for 1-shot while $\approx 8-32\%$ on clean data and $\approx 31-40\%$ on adversarial data for 5-shot settings), over existing state-of-the-art methods. 

We obtain similar observations on a large scale Mini-ImageNet, which is another popular benchmark dataset. Overall, we significantly outperform the existing state-of-the-art robust few-shot methods on both clean and adversarial accuracy. Thus, it clearly highlights the utility of our frequency-regularized pretraining and finetuning. 
Note that in contrast to existing few-shot robustness techniques, we don’t generate adversarial samples during pre-training or fine-tuning. This allows us to significantly reduce the training time while simultaneously outperforming them in both clean and robust accuracies (refer Fig.~\ref{fig:motivation}).

\subsection{Varying Support and Query set size}

\begin{table}[hbp]
\caption{Analyzing the sensitivity of our proposed approach towards variations in amount of per-class samples in the support set for a $5$-way setting on CIFAR-FS dataset.
}
\label{tab:support_set_vary}
\centering
\scalebox{0.9}{
\begingroup
\setlength{\tabcolsep}{0.12cm} 
\renewcommand{\arraystretch}{1.3}
\begin{tabular}{|c|cccc|}
\hline
\multirow{3}{*}{\textbf{\begin{tabular}[c]{@{}c@{}}Support Set \\ (Mix, Max, \\ Class Distribution)\end{tabular}}} &
  \multicolumn{4}{c|}{\textbf{Technique}} \\ \cline{2-5} 
 &
  \multicolumn{2}{c|}{\textbf{Ours Transductive}} &
  \multicolumn{2}{c|}{\textbf{Ours  Non-Transductive}} \\ \cline{2-5} 
 &
  \multicolumn{1}{c|}{Clean} &
  \multicolumn{1}{c|}{PGD} &
  \multicolumn{1}{c|}{Clean} &
  PGD \\ \hline
\begin{tabular}[c]{@{}c@{}}1, 5, {[}5 5 5 5 5{]}\\ (balanced)\end{tabular} &
  \multicolumn{1}{c|}{79.79 (0.55)} &
  \multicolumn{1}{c|}{75.93 (0.61)} &
  \multicolumn{1}{c|}{73.91 (0.55)} &
  61.82 (0.65) \\ \hline 
\begin{tabular}[c]{@{}c@{}}1, 5, {[}4 5 3 5 5{]}\\ (unbalanced)\end{tabular} &
  \multicolumn{1}{c|}{78.99 (0.54)} &
  \multicolumn{1}{c|}{74.81 (0.61)} &
  \multicolumn{1}{c|}{73.64 (0.55)} &
  61.46 (0.65) \\ \hline \hline
\begin{tabular}[c]{@{}c@{}}1, 10, {[}10 10 10 10 10{]}\\ (balanced)\end{tabular} &
  \multicolumn{1}{c|}{81.85 (0.51)} &
  \multicolumn{1}{c|}{78.28 (0.58)} &
  \multicolumn{1}{c|}{77.98 (0.51)} &
  66.94 (0.62) \\ \hline
\begin{tabular}[c]{@{}c@{}}1, 10, {[}7 4 8 5 7{]}\\ (unbalanced)\end{tabular} &
  \multicolumn{1}{c|}{82.17 (0.50)} &
  \multicolumn{1}{c|}{78.20 (0.57)} &
  \multicolumn{1}{c|}{77.67 (0.52)} &
  66.57 (0.62) \\ \hline \hline
\begin{tabular}[c]{@{}c@{}}1, 15, {[}15 15 15 15 15{]}\\ (balanced)\end{tabular} &
  \multicolumn{1}{c|}{83.38 (0.50)} &
  \multicolumn{1}{c|}{79.63 (0.57)} &
  \multicolumn{1}{c|}{79.92 (0.50)} &
  69.17 (0.61) \\ \hline
\begin{tabular}[c]{@{}c@{}}1, 15, {[} 7  4 13 15 11{]}\\ (unbalanced)\end{tabular} &
  \multicolumn{1}{c|}{83.10 (0.50)} &
  \multicolumn{1}{c|}{79.31 (0.57)} &
  \multicolumn{1}{c|}{79.11 (0.50)} &
  68.49 (0.60) \\ \hline
\end{tabular}
\endgroup
}
\end{table}
In this section, we evaluate our proposed method in a more general setting where we also vary the amount of samples per class in the support set and the quantity of query set during inference. This setting is in contrast with the traditional setup (for a $k$-way $n$-shot, fixed $n$ across the $k$ classes and fixed query set size) that are typically followed in robust few-shot domain.

We randomly select the support set size of each class instead of fixing it to a class-balanced value like 1 or 5 (as used in the main paper). For instance, in a $k$-way setting we sample $k$ random numbers between [1, $max\_shot\_value$], constraining each class to have minimum $1$ support set sample and maximum $max\_shot\_value$. We perform experiments on CIFAR-FS dataset and fix the number of classes i.e. $k$ as $5$ and query set size as $75$ ($15$ query samples per class). In Table~\ref{tab:support_set_vary}, we observe that there is only a minor drop for both clean and robust accuracy on our proposed method (for both transductive and non-transductive setups) in the random sampling (unbalanced) setting compared to the balanced setting. The trend is consistent as we vary the $max\_shot\_value$ from $5$ to $15$.

In Table~\ref{tab:query_set_size_vary} we vary the quantity of the query set from $5$ samples per class (i.e. query set size = $25$) to $25$ samples per class (i.e. query set size = $125$) on CIFAR-FS for $5$-way $1$-shot setting. We observe that our proposed method (both transductive and non-transductive finetuning) performs decently well as the performance variation in both clean and robust accuracy is mild, when the query set size is scaled up.


\begin{table}[htp]
\caption{Investigating the effect of varying the size of query set from $5$ samples per class to $25$ samples per class for $5$-way $1$-shot setting on CIFAR-FS dataset.}
\label{tab:query_set_size_vary}
\centering
\scalebox{0.9}{
\begingroup
\setlength{\tabcolsep}{0.18cm} 
\renewcommand{\arraystretch}{1.3}
\begin{tabular}{|c|cccc|}
\hline
\multirow{2}{*}{\textbf{Query Set Size}} & \multicolumn{4}{c|}{Technique}                                                                                           \\ \cline{2-5} 
   ($n_q$)                            & \multicolumn{2}{c|}{\textbf{Ours Transductive}}                                 & \multicolumn{2}{c|}{\textbf{Ours Non-Transductive}}        \\ \cline{2-5} 
                               & \multicolumn{1}{c|}{Clean}        & \multicolumn{1}{c|}{PGD}          & \multicolumn{1}{c|}{Clean}        & PGD          \\ \hline
25                              & \multicolumn{1}{c|}{62.95 (0.84)}             & \multicolumn{1}{c|}{60.47 (0.86)}             & \multicolumn{1}{c|}{55.22 (0.80)}             &   41.64 (0.79)           \\ \hline
50                              & \multicolumn{1}{c|}{63.24 (0.74)}             & \multicolumn{1}{c|}{59.30 (0.78)}             & \multicolumn{1}{c|}{55.10 (0.70)}             &       41.77 (0.70)       \\ \hline
75                              & \multicolumn{1}{c|}{65.03 (0.72)} & \multicolumn{1}{c|}{60.55 (0.76)} & \multicolumn{1}{c|}{55.00 (0.69)} & 41.55 (0.70) \\ \hline
100                             & \multicolumn{1}{c|}{64.30 (0.70)} & \multicolumn{1}{c|}{58.90 (0.75)} & \multicolumn{1}{c|}{55.03 (0.68)} & 41.63 (0.68) \\ \hline
125                             & \multicolumn{1}{c|}{64.94 (0.70)} & \multicolumn{1}{c|}{59.08 (0.74)} & \multicolumn{1}{c|}{54.98 (0.66)} & 41.55 (0.67) \\ \hline
\end{tabular}
\endgroup
}
\end{table}
\subsection{Sensitivity Analysis: $\lambda$ value in weighted-sampling}
\label{subsection:lambda_sens}
\begin{figure}[htp]
\centering
\centerline{\includegraphics[width=0.5\textwidth]{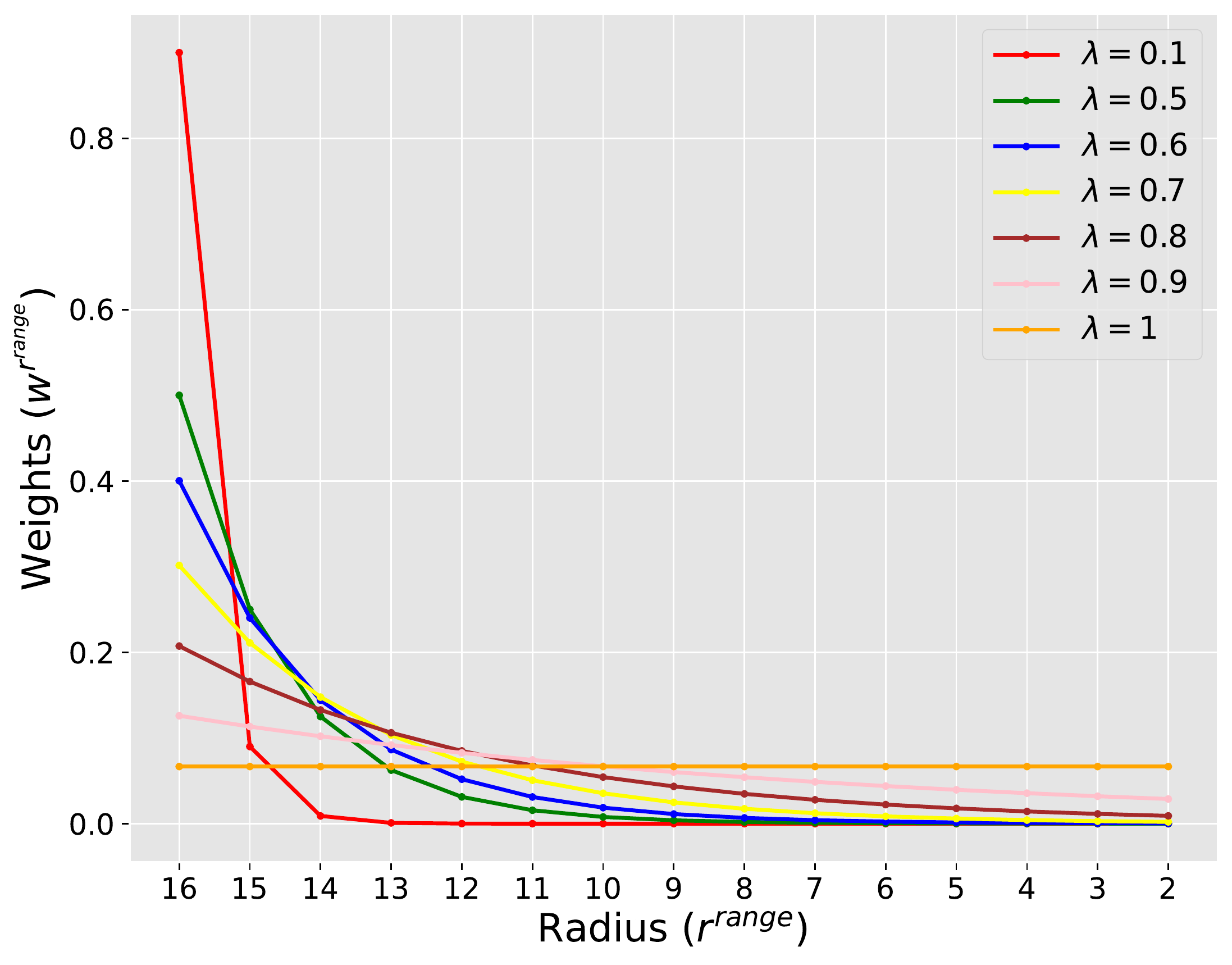}}
\caption{\small{Demonstrating the change in the long-tail distribution when $\lambda$ is varied. The distribution approximates to sampling from a fixed radius for very small values of $\lambda$, whereas it becomes uniform distribution for very high values of $\lambda$ (close to $1$).}} 
\label{fig:long_tail}
\end{figure}
In the progressive learning scheme (eq.~\ref{eq6}), we use a distribution (i.e. Long Tailed) over $r^{range}$ to select the radius stochastically. As the distribution depends on input parameter $\lambda$, it becomes crucial to study the effect of $\lambda$ on the distribution. Fig.~\ref{fig:long_tail} shows the changes in the distribution on varying the value of parameter $\lambda$. It can be observed that for the small values (i.e. $0.1$) of $\lambda$, the weights are concentrated over a very small range in $r^{range}$ approximating it to as sampling from a fixed radius. Whereas for a higher value of $\lambda$ (close to $1$), the weights become uniform throughout the $r^{range}$. Earlier in our experiments (section~\ref{subsection: PL}), we showed that for both the extremum condition, i.e., using a fixed radius and the uniform distribution over $r^{range}$  gives poor performance. Hence we expect to obtain the optimal results for some intermediate values of $\lambda$. 

\begin{figure}[htp]
\centering
\centerline{\includegraphics[width=0.5\textwidth]{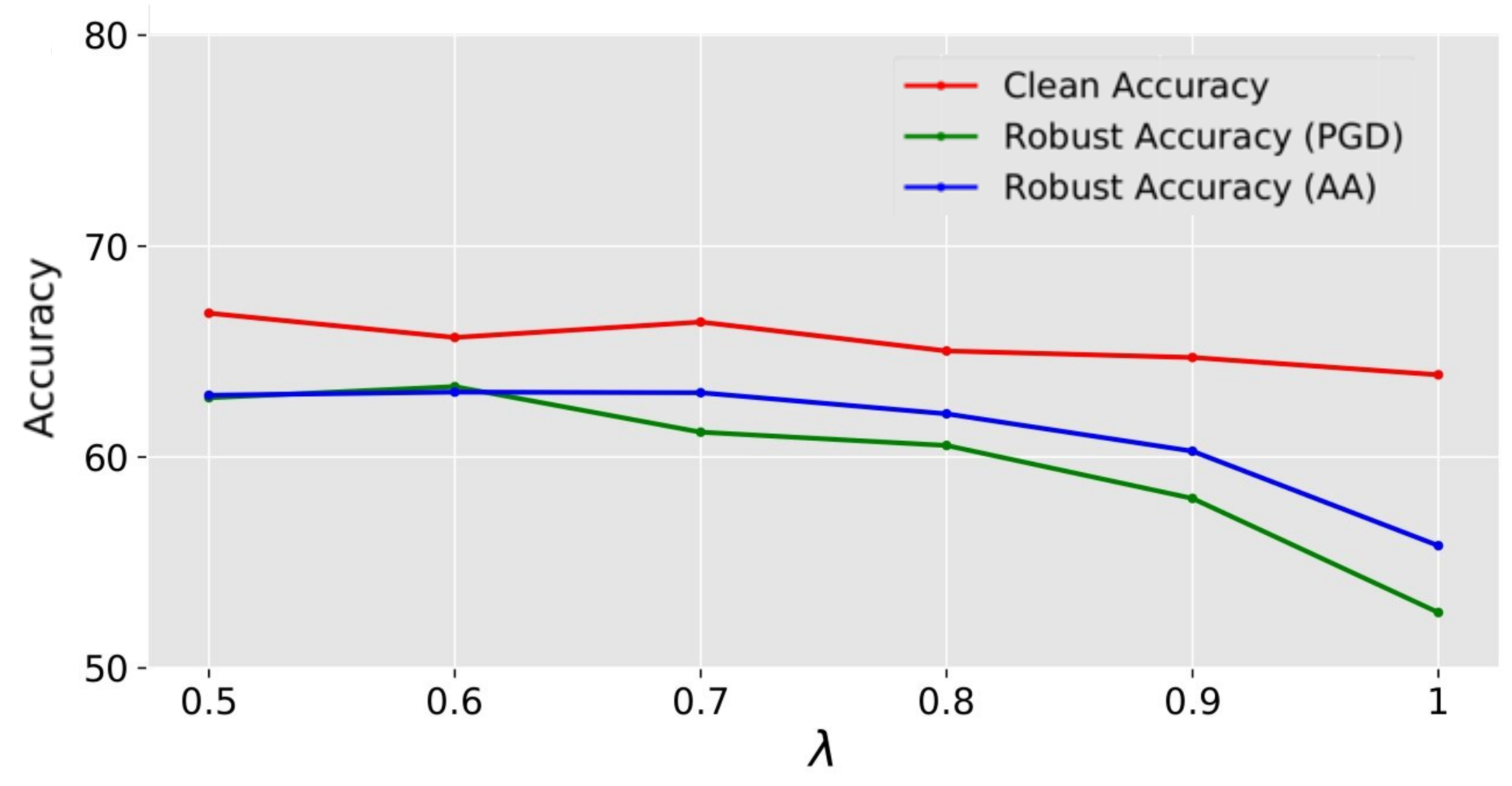}}
\caption{\small{Variation in performance of our proposed approach with respect to $\lambda$ on CIFAR-FS. $\lambda$ is varied for different intermediate values starting from $0.5$. Both the clean and robust performance is not much sensitive when $\lambda$ takes values between $0.5$ to $0.8$. PGD and AA refers to the adversarial attacks using Projected Gradient Descent and Auto Attack respectively.}} 
\label{fig:lambda_result}
\end{figure}
Thus to verify it, in this section, we perform experiments by taking different intermediate values of $\lambda$ starting from $0.5$. The obtained results are shown in Fig~\ref{fig:lambda_result}. It can be observed that the clean accuracy of the model over different values of $\lambda$ has minimal variation. In the case of Robust Accuracy, there is not much variation in range [$0.5$-$0.8$]. However, it starts dropping significantly as the value of $\lambda$ increases after $0.8$. This validates our initial intuition where we expected poor performance near $\lambda$ close to $1$ (approximating to uniform distribution).

Through this analysis, we found selecting $\lambda$ from an intermediate range i.e. [$0.5$-$0.8$] can provide decent overall performance, as the performance is not very sensitive to $\lambda$ in this range. Specifically, even though performance corresponding to $\lambda = 0.6$ seems to be the overall-best, $\lambda = 0.8$ also gave competitive performance in all our experiments
, which further confirms that the performance is not sensitive to the choice of $\lambda$ in the intermediate range.  
\subsection{t-SNE Visualizations}

\begin{figure*}[htp]
\centering
\centerline{\includegraphics[width=1.1\textwidth]{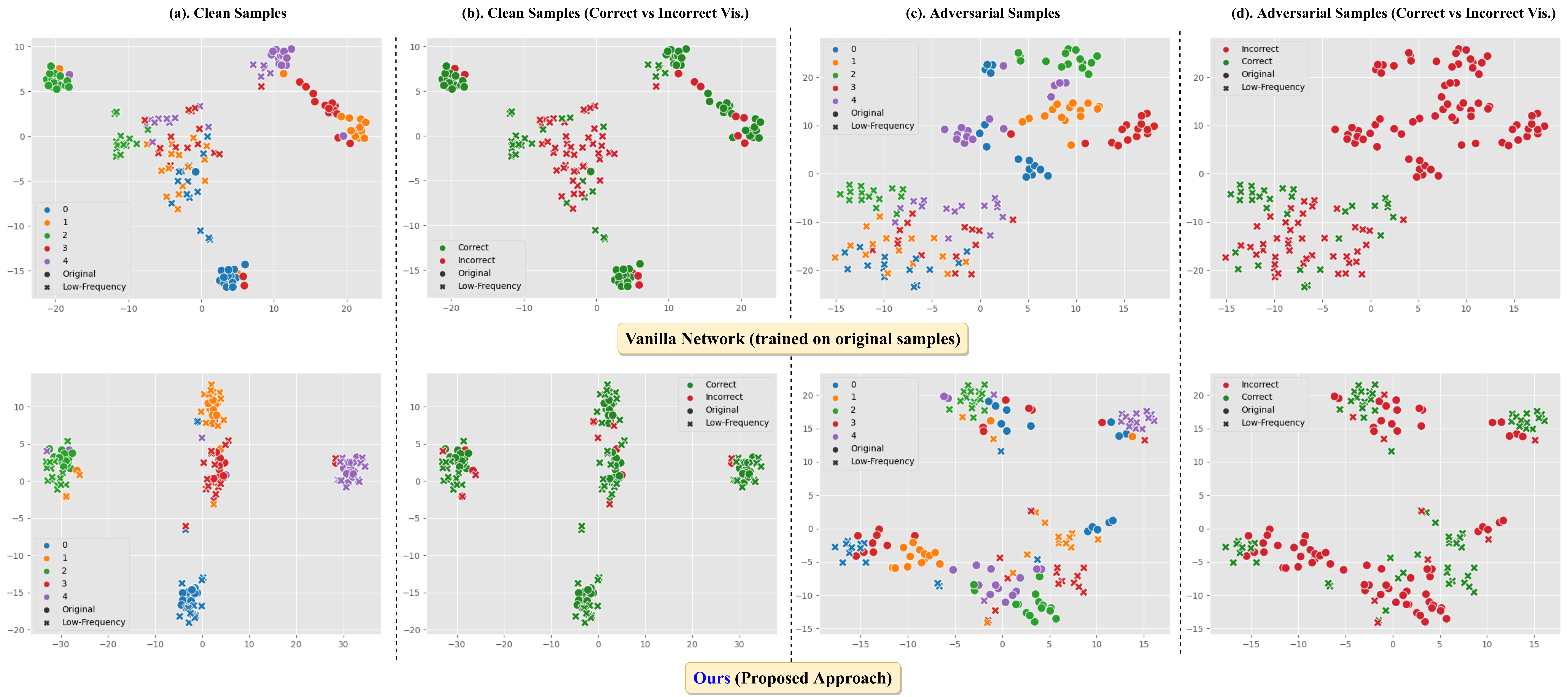}}
\caption{\small{t-SNE visualizations for the features of clean original and their corresponding low-frequency samples as well as the features of adversarially perturbed original and corresponding low-frequency samples that are obtained by the vanilla model and our proposed approach. We compare our approach and vanilla baseline by demonstrating/visualizing (a) feature representations marked on the basis of the predicted class 
and (b) the predictions on the features compared with the ground truth (marked correct or incorrect).}} 
\label{fig:tsne-clean}
\end{figure*}

In this section, we qualitatively analyze the performance of our proposed approach by visualizing the feature representations obtained at the pre-final layer of the model. We visualize the feature representations of the query samples using t-SNE algorithm \cite{Maaten2008VisualizingDU} for a 5-way 5-shot setting on CIFAR-FS dataset. The column (a) and (c) in Figure~\ref{fig:tsne-clean} 
represents the class-wise (i.e. with respect to the predicted class) feature visualization, while column (b) and (d) indicates whether the sample was correctly classified or not by the model.

We first visualize the features of clean original and their corresponding low-frequency samples in (a) and (b). The first row represents the visualizations corresponding to the vanilla network (i.e. trained transductively~\cite{Dhillon2020A} on original samples), which served as a baseline. We observe that although the vanilla network is able to correctly classify the original samples (represented via symbol ‘o’) (Figure~\ref{fig:tsne-clean} row-1 (b)), it is not able to achieve similar performance on the corresponding low-frequency samples (represented via symbol `$\times$'). This behavior can be primarily attributed to the low-discriminative power (and consequently poor features learned) on low-frequency samples by the vanilla networks (Figure~\ref{fig:tsne-clean} row-1 (a)). Next, in the second row we visualize the feature representations of our proposed model on the clean data. We observe that we achieve significantly better performance on the low-frequency samples (Figure~\ref{fig:tsne-clean} row-2 (b)) without sacrificing performance on original samples. As we will see in the subsequent paragraph, this improvement in discriminability on low-frequency samples is key to achieving significant boost in robustness accuracy without much drop on the clean accuracy.

We also visualize the feature representations of adversarially perturbed samples and their corresponding low-frequency counterparts. We note that the vanilla model is highly vulnerable to adversarial attacks as it misclassified all the adversarially perturbed ‘original’ samples (refer Figure~\ref{fig:tsne-clean} row-1 (d)). However, this trend doesn’t hold for adversarially perturbed low-frequency samples as the vanilla network is able to predict a fraction of them correctly. Thus, some non-trivial robustness can be achieved for the vanilla model by simply masking out the high-frequency components, however, the robustness-performance is limited by low-discriminative power on the low-frequency samples. Finally, we visualize the feature representations learned using our proposed approach and note that we are able to correctly classify the majority of low-frequency representation of adversarially perturbed  samples (refer Figure~\ref{fig:tsne-clean} row-2 (d)). This observation is in line with our empirical results described earlier, 
where we noticed a significant boost in the robustness accuracy through our proposed approach compared to the vanilla model.

\section{Discussion and Conclusion}
In this work, we present for the first time a non-meta learning method for robust few-shot learning that achieves state-of-the-art robust accuracy (without compromising much on clean data accuracy) on multiple benchmarks with marginal computational overhead. We primarily achieve this by learning rich discriminative features on low-frequency data in a self-distillation setup during pretraining. 
In the finetuning stage, the student network further benefits by applying additional cosine similarity loss. We also qualitatively verify it by analyzing our proposed approach through visualization of the t-SNE plots of the feature representations of clean and adversarial query samples with their corresponding low-frequency counterparts. 

Our PL module automates the process of selecting the radius (distribution) in an end-to-end data-driven manner. This is especially important as selecting a fixed radius (for e.g. $r=2$) might work well for certain dataset (for instance CIFAR-FS) but can yield suboptimal performance on others (for e.g. Mini-ImageNet). On the contrary, our PL module learns the weight distribution (i.e. allocation of weights to radii) in a data-driven manner during the pretraining. This ensures that the knowledge acquired (regarding the weight distribution) during pre-training is suitable \& specific to the dataset-domain and model architecture. 
We also observed that using weighted-ensemble for evaluation usually provided better clean and robust accuracy, compared to evaluating on a peak radius (radius corresponding to highest weight) as it appropriately weighs the logit contribution from each radius. 
As evidenced by experimental results, our method leads to improvement in robust accuracy without 
compromising much on clean accuracy and is trained with significantly lesser computationally overhead as compared to robust meta-learning methods. 

We hope the simplicity yet strong performance of our method would serve as a compelling baseline to robust few-shot learning methods and urge research community to explore more non-meta learning and adversarial-sample free robustness methods. 

\section{Acknowledgments}
\noindent This work is partially supported by a Young Scientist Research Award (Sanction no. 59/20/11/2020-BRNS) to Anirban Chakraborty from DAE-BRNS, India.

\bibliographystyle{IEEEtran}
\bibliography{references}

\begin{thebibliography}{10}
\providecommand{\url}[1]{#1}
\csname url@samestyle\endcsname
\providecommand{\newblock}{\relax}
\providecommand{\bibinfo}[2]{#2}
\providecommand{\BIBentrySTDinterwordspacing}{\spaceskip=0pt\relax}
\providecommand{\BIBentryALTinterwordstretchfactor}{4}
\providecommand{\BIBentryALTinterwordspacing}{\spaceskip=\fontdimen2\font plus
\BIBentryALTinterwordstretchfactor\fontdimen3\font minus
  \fontdimen4\font\relax}
\providecommand{\BIBforeignlanguage}[2]{{%
\expandafter\ifx\csname l@#1\endcsname\relax
\typeout{** WARNING: IEEEtran.bst: No hyphenation pattern has been}%
\typeout{** loaded for the language `#1'. Using the pattern for}%
\typeout{** the default language instead.}%
\else
\language=\csname l@#1\endcsname
\fi
#2}}
\providecommand{\BIBdecl}{\relax}
\BIBdecl

\bibitem{szegedy2013intriguing}
\BIBentryALTinterwordspacing
C.~Szegedy, W.~Zaremba, I.~Sutskever, J.~Bruna, D.~Erhan, I.~J. Goodfellow, and
  R.~Fergus, ``Intriguing properties of neural networks,'' in
  \emph{International Conference on Learning Representations, {ICLR} 2014,
  Conference Track Proceedings}, 2014. [Online]. Available:
  \url{http://arxiv.org/abs/1312.6199}
\BIBentrySTDinterwordspacing

\bibitem{goodfellow2014explaining}
\BIBentryALTinterwordspacing
I.~J. Goodfellow, J.~Shlens, and C.~Szegedy, ``Explaining and harnessing
  adversarial examples,'' in \emph{International Conference on Learning
  Representations, {ICLR} 2015, Conference Track Proceedings}, 2015. [Online].
  Available: \url{http://arxiv.org/abs/1412.6572}
\BIBentrySTDinterwordspacing

\bibitem{yang2020finding}
J.~Yang, A.~Boloor, A.~Chakrabarti, X.~Zhang, and Y.~Vorobeychik, ``Finding
  physical adversarial examples for autonomous driving with fast and
  differentiable image compositing,'' 2021.

\bibitem{fei2020adversarial}
J.~Fei, Z.~Xia, P.~Yu, and F.~Xiao, ``Adversarial attacks on fingerprint
  liveness detection,'' \emph{EURASIP Journal on Image and Video Processing},
  vol. 2020, no.~1, pp. 1--11, 2020.

\bibitem{madry2017towards}
A.~Madry, A.~Makelov, L.~Schmidt, D.~Tsipras, and A.~Vladu, ``Towards deep
  learning models resistant to adversarial attacks,'' in \emph{International
  Conference on Learning Representations}, 2018.

\bibitem{carmon2019unlabeled}
Y.~Carmon, A.~Raghunathan, L.~Schmidt, P.~Liang, and J.~Duchi, ``Unlabeled data
  improves adversarial robustness,'' in \emph{Advances in Neural Information
  Processing Systems (NeurIPS)}, 2019.

\bibitem{alayrac2019labels}
J.-B. Alayrac, J.~Uesato, P.-S. Huang, A.~Fawzi, R.~Stanforth, and P.~Kohli,
  ``Are labels required for improving adversarial robustness?'' \emph{Advances
  in Neural Information Processing Systems}, vol.~32, 2019.

\bibitem{goldblum2020adversarially}
M.~Goldblum, L.~Fowl, and T.~Goldstein, ``Adversarially robust few-shot
  learning: A meta-learning approach,'' \emph{Advances in Neural Information
  Processing Systems}, vol.~33, 2020.

\bibitem{snell2017prototypical}
J.~Snell, K.~Swersky, and R.~Zemel, ``Prototypical networks for few-shot
  learning,'' \emph{Advances in neural information processing systems},
  vol.~30, 2017.

\bibitem{bertinetto2018metalearning}
L.~Bertinetto, J.~F. Henriques, P.~Torr, and A.~Vedaldi, ``Meta-learning with
  differentiable closed-form solvers,'' in \emph{International Conference on
  Learning Representations}, 2019.

\bibitem{lee2019meta}
K.~Lee, S.~Maji, A.~Ravichandran, and S.~Soatto, ``Meta-learning with
  differentiable convex optimization,'' in \emph{Proceedings of the IEEE/CVF
  Conference on Computer Vision and Pattern Recognition}, 2019.

\bibitem{yin2018adversarial}
\BIBentryALTinterwordspacing
C.~Yin, J.~Tang, Z.~Xu, and Y.~Wang, ``Adversarial meta-learning,'' 2021.
  [Online]. Available: \url{https://openreview.net/forum?id=Z_3x5eFk1l-}
\BIBentrySTDinterwordspacing

\bibitem{sung2018learning}
F.~Sung, Y.~Yang, L.~Zhang, T.~Xiang, P.~H. Torr, and T.~M. Hospedales,
  ``Learning to compare: Relation network for few-shot learning,'' in
  \emph{Proceedings of the IEEE conference on computer vision and pattern
  recognition}, 2018, pp. 1199--1208.

\bibitem{vinyals2016matching}
O.~Vinyals, C.~Blundell, T.~Lillicrap, D.~Wierstra \emph{et~al.}, ``Matching
  networks for one shot learning,'' \emph{Advances in neural information
  processing systems}, vol.~29, pp. 3630--3638, 2016.

\bibitem{finn2017model}
C.~Finn, P.~Abbeel, and S.~Levine, ``Model-agnostic meta-learning for fast
  adaptation of deep networks,'' in \emph{International Conference on Machine
  Learning}.\hskip 1em plus 0.5em minus 0.4em\relax PMLR, 2017, pp. 1126--1135.

\bibitem{wang2020high}
H.~Wang, X.~Wu, Z.~Huang, and E.~P. Xing, ``High-frequency component helps
  explain the generalization of convolutional neural networks,'' in
  \emph{Proceedings of the IEEE/CVF Conference on Computer Vision and Pattern
  Recognition}, 2020, pp. 8684--8694.

\bibitem{wang2020towards}
Z.~Wang, Y.~Yang, A.~Shrivastava, V.~Rawal, and Z.~Ding, ``Towards
  frequency-based explanation for robust cnn,'' \emph{arXiv preprint
  arXiv:2005.03141}, 2020.

\bibitem{Dhillon2020A}
G.~S. Dhillon, P.~Chaudhari, A.~Ravichandran, and S.~Soatto, ``A baseline for
  few-shot image classification,'' in \emph{International Conference on
  Learning Representations}, 2020.

\bibitem{zhang2019theoretically}
H.~Zhang, Y.~Yu, J.~Jiao, E.~Xing, L.~El~Ghaoui, and M.~Jordan, ``Theoretically
  principled trade-off between robustness and accuracy,'' in
  \emph{International conference on machine learning}.\hskip 1em plus 0.5em
  minus 0.4em\relax PMLR, 2019, pp. 7472--7482.

\bibitem{zi2021revisiting}
B.~Zi, S.~Zhao, X.~Ma, and Y.-G. Jiang, ``Revisiting adversarial robustness
  distillation: Robust soft labels make student better,'' in \emph{Proceedings
  of the IEEE/CVF International Conference on Computer Vision}, 2021, pp.
  16\,443--16\,452.

\bibitem{chan2019jacobian}
A.~Chan, Y.~Tay, Y.~S. Ong, and J.~Fu, ``Jacobian adversarially regularized
  networks for robustness,'' in \emph{International Conference on Learning
  Representations}, 2020.

\bibitem{Addepalli_2020_CVPR}
S.~Addepalli, V.~B.S., A.~Baburaj, G.~Sriramanan, and R.~V. Babu, ``Towards
  achieving adversarial robustness by enforcing feature consistency across bit
  planes,'' in \emph{Proceedings of the IEEE/CVF Conference on Computer Vision
  and Pattern Recognition (CVPR)}, June 2020.

\bibitem{chen2019improving}
H.-Y. Chen, J.-H. Liang, S.-C. Chang, J.-Y. Pan, Y.-T. Chen, W.~Wei, and D.-C.
  Juan, ``Improving adversarial robustness via guided complement entropy,'' in
  \emph{Proceedings of the IEEE/CVF International Conference on Computer
  Vision}, 2019, pp. 4881--4889.

\bibitem{heattack2017}
W.~He, J.~Wei, X.~Chen, N.~Carlini, and D.~Song, ``Adversarial example
  defenses: ensembles of weak defenses are not strong,'' in \emph{Proceedings
  of the 11th USENIX Conference on Offensive Technologies}, 2017, pp. 15--15.

\bibitem{bengio2012deep}
Y.~Bengio, ``Deep learning of representations for unsupervised and transfer
  learning,'' in \emph{Proceedings of ICML workshop on unsupervised and
  transfer learning}.\hskip 1em plus 0.5em minus 0.4em\relax JMLR Workshop and
  Conference Proceedings, 2012, pp. 17--36.

\bibitem{chen2019closerfewshot}
W.-Y. Chen, Y.-C. Liu, Z.~Kira, Y.-C. Wang, and J.-B. Huang, ``A closer look at
  few-shot classification,'' in \emph{International Conference on Learning
  Representations}, 2019.

\bibitem{bertinetto2018meta}
L.~Bertinetto, J.~F. Henriques, P.~Torr, and A.~Vedaldi, ``Meta-learning with
  differentiable closed-form solvers,'' in \emph{International Conference on
  Learning Representations}, 2019.

\bibitem{liu2021longterm}
F.~Liu, S.~Zhao, X.~Dai, and B.~Xiao, ``Long-term cross adversarial training: A
  robust meta-learning method for few-shot classification tasks,'' in
  \emph{ICML 2021 Workshop on Adversarial Machine Learning}, 2021.

\bibitem{Li2022robust}
Z.~Li, J.~O. Caro, E.~Rusak, W.~Brendel, M.~Bethge, F.~Anselmi, A.~B. Patel,
  A.~S. Tolias, and X.~Pitkow, ``Robust deep learning object recognition models
  rely on low frequency information in natural images,'' \emph{bioRxiv}, 2022.

\bibitem{he2016deep}
K.~He, X.~Zhang, S.~Ren, and J.~Sun, ``Deep residual learning for image
  recognition,'' in \emph{Proceedings of the IEEE conference on computer vision
  and pattern recognition}, 2016, pp. 770--778.

\bibitem{Maaten2008VisualizingDU}
L.~van~der Maaten and G.~E. Hinton, ``Visualizing data using t-sne,''
  \emph{Journal of Machine Learning Research}, vol.~9, pp. 2579--2605, 2008.

\bibitem{JMLR:v13:gretton12a}
\BIBentryALTinterwordspacing
A.~Gretton, K.~M. Borgwardt, M.~J. Rasch, B.~Sch{{\"o}}lkopf, and A.~Smola, ``A
  kernel two-sample test,'' \emph{Journal of Machine Learning Research},
  vol.~13, no.~25, pp. 723--773, 2012. [Online]. Available:
  \url{http://jmlr.org/papers/v13/gretton12a.html}
\BIBentrySTDinterwordspacing

\bibitem{kullback1951information}
S.~Kullback and R.~A. Leibler, ``On information and sufficiency,'' \emph{The
  annals of mathematical statistics}, vol.~22, no.~1, pp. 79--86, 1951.

\bibitem{zagoruyko2016wide}
\BIBentryALTinterwordspacing
S.~Zagoruyko and N.~Komodakis, ``Wide residual networks,'' in \emph{Proceedings
  of the British Machine Vision Conference (BMVC)}, E.~R.~H. Richard C.~Wilson
  and W.~A.~P. Smith, Eds.\hskip 1em plus 0.5em minus 0.4em\relax BMVA Press,
  September 2016, pp. 87.1--87.12. [Online]. Available:
  \url{https://dx.doi.org/10.5244/C.30.87}
\BIBentrySTDinterwordspacing

\end{thebibliography}

\newpage
\onecolumn
{\centering{\LARGE \textbf{\textit{Supplementary for\\} \vspace{2ex}``Robust Few-shot Learning Without Using any Adversarial Samples"} \par}\vspace{5ex}
	
	}

\setcounter{section}{0}
\setcounter{table}{0}
\setcounter{figure}{0}
\vspace{8pt}
\hrule
\vspace{18pt}

This supplementary document is organized as follows:
\vspace{0.1in}
\begin{itemize}
\item Additional Ablations
\begin{itemize}
\item Different losses for Frequency Regularization
\item Different Backbone Architectures
\item Finetuning with Cosine Similarity loss ($L_{cs}$) on either support, query or both 
\end{itemize}
\item Sensitivity Analysis 
\begin{itemize}
\item Radius Range ($r^{range}$) in Progressive Learning
\end{itemize}
\item Additional Training details
\item Qualitative Demonstration
\item Pseudocode of Proposed Algorithm
\end{itemize}

\vspace{0.1in}
\section{Additional Ablations}
\vspace{0.2in}
\subsection{Different losses for Frequency Regularization}
To enhance the performance of $S$ on low-frequency samples, we use the low-frequency regularization $L_{cs}$  during pre-training ((eq.$ 5.$) of the main paper) and fine-tuning ((eq.$ 7.$) of the main paper) . This helps in matching the high-level feature representation of low-frequency samples to the original one, making the feature discriminative. Apart from cosine similarity, other popular techniques to align these feature representations are MMD ~\cite{JMLR:v13:gretton12a}, KL divergence \cite{kullback1951information}. MMD Loss tries to match the moments of two distributions by minimizing the kernel distance between them. KL divergence loss, on the other hand, reduces the shift between the two probability distributions in the perspective of information theory. In this section, we investigate the impact on performance for the above choices.

\begin{table}[h]
\caption{Ablation on different choices of loss functions that can be used for frequency regularization.}
\centering
\scalebox{0.90}{
\begingroup
\setlength{\tabcolsep}{0.18cm} 
\renewcommand{\arraystretch}{1.6}
\begin{tabular}{|c|c|ccc|ccc|}
\hline
\multicolumn{1}{|c|}{Setup} & \multicolumn{1}{c|}{Method} &
  \multicolumn{1}{c|}{Clean} &
  \multicolumn{1}{c|}{PGD} &
  \begin{tabular}[c]{@{}c@{}}Auto Attack\end{tabular} \\ \hline \hline
\multirow{4}{*}{1 - SHOT} & Baseline &
  \multicolumn{1}{c|}{68.40 (0.71)} &
  \multicolumn{1}{c|}{0.01 (0.01)} &
  0.00 (0.00)  \\ \cline{2-5} \cline{2-5}
& \begin{tabular}[c]{@{}c@{}}Cosine Similarity\end{tabular} &
  \multicolumn{1}{c|}{65.03 (0.72)} &
  \multicolumn{1}{c|}{60.55 (0.76)} &
  62.05 (0.74)  \\ \cline{2-5}
& \begin{tabular}[c]{@{}c@{}}KL Divergence\end{tabular} &
  \multicolumn{1}{c|}{64.73 (0.73)} &
  \multicolumn{1}{c|}{61.09 (0.77)} &
  \multicolumn{1}{c|}{62.31 (0.76)}
   \\ \cline{2-5}
& MMD                             & \multicolumn{1}{c|}{67.94 (0.72)} & \multicolumn{1}{c|}{60.10 (0.81)} & \multicolumn{1}{c|}{62.59 (0.79)}  \\ \hline

\multirow{4}{*}{5 - SHOT} & Baseline &
  \multicolumn{1}{c|}{81.81 (0.52)} &
  \multicolumn{1}{c|}{0.07 (0.02)} &
  0.00 (0.00) \\ \cline{2-5} \cline{2-5}
& \begin{tabular}[c]{@{}c@{}}Cosine Similarity\end{tabular} &
  \multicolumn{1}{c|}{79.79 (0.55)} &
  \multicolumn{1}{c|}{75.93 (0.61)} &
  77.08 (0.59) \\ \cline{2-5}
& \begin{tabular}[c]{@{}c@{}}KL Divergence\end{tabular} &
  \multicolumn{1}{c|}{78.87 (0.57)} &
  \multicolumn{1}{c|}{75.33 (0.63)} &
  \multicolumn{1}{c|}{76.42 (0.61)}
   \\ \cline{2-5}
& MMD                             & \multicolumn{1}{c|}{79.77 (0.56)} & \multicolumn{1}{c|}{72.44 (0.67)} & \multicolumn{1}{c|}{76.92 (0.57)}  \\ \hline

\end{tabular}
\endgroup
}
\label{tab:table3}
\end{table}

We notice (refer Table-\ref{tab:table3}) that Cosine Sim. performs (similar or) better than both KL divergence and MMD on the 5-SHOT setting.  Whereas, MMD performs slightly better in clean-accuracy and similar in the PGD and AutoAttack accuracies, compared to Cosine Sim. and KL-divergence in the 1-SHOT setting. Owing to the ease-of-use, (relatively) lower time consumption (especially compared to MMD which results in a quadratic computation complexity \cite{JMLR:v13:gretton12a}) leading to cheaper computational expense and overall (similar or) better performance on robust accuracies in both 5-SHOT and 1-SHOT settings, we opt for Cosine Sim. loss for performing frequency regularization.
\newpage
\subsection{Different Backbone Architectures}
In our main paper, throughout all the experiments, we used Resnet-$12$ \cite{he2016deep} as backbone architecture for both the teacher $T$ and student $S$ as described in experiments section (Section V in the main paper). Due to the availability of a wide variety of DNNs with varying network capacities in literature, it becomes vital to evaluate the performance of the proposed approach across such different choices of backbone architectures. Hence in this section, we investigate the performance of our approach when built upon some other popular architecture used in practice. Specifically, we repeat our experiments using  WRN-$28$-$10$ (Wide Resnet with width $28$ and depth $10$) \cite{zagoruyko2016wide} and conv $(64)_{\times4}$ (CNN with $4$ layers and $64^{k}$ channels in the $k^{th}$ layer) \cite{vinyals2016matching} on CIFAR-FS \cite{bertinetto2018meta} for both $1$-shot and $5$-shot settings and report their results in Table \ref{tab:table1}. 

\begin{table*}[htp]
\caption{Performance of our proposed method across different backbone architectures on CIFAR-FS dataset. We observe significant gains in adversarial accuracy even across different backbone architectures. The performance on 5-shot setting is even better than 1-shot. We are able to retain clean performance on Conv-64 while more gains in adversarial performance on Resnet-12 and WRN-28-10.}
\centering
\scalebox{0.90}{
\begingroup
\setlength{\tabcolsep}{0.18cm} 
\renewcommand{\arraystretch}{1.6}
\begin{tabular}{|c|ccc|ccc|}
\hline
\multirow{2}{*}{\textbf{Setup}} &
  \multicolumn{3}{c|}{\textbf{1 - SHOT}} &
  \multicolumn{3}{c|}{\textbf{5 - SHOT}} \\ \cline{2-7} 
 &
  \multicolumn{1}{c|}{Clean} &
  \multicolumn{1}{c|}{PGD} &
  \begin{tabular}[c]{@{}c@{}}Auto Attack\end{tabular} &
  \multicolumn{1}{c|}{Clean} &
  \multicolumn{1}{c|}{PGD} &
  \begin{tabular}[c]{@{}c@{}}Auto Attack\end{tabular} \\ \hline \hline
\begin{tabular}[c]{@{}c@{}}Conv-64 (Baseline)\end{tabular} &
  \multicolumn{1}{c|}{55.75 (0.75)} &
  \multicolumn{1}{c|}{0.00 (0.00)} &
  0.00 (0.00) &
  \multicolumn{1}{c|}{73.37 (0.56)} &
  \multicolumn{1}{c|}{0.00 (0.00)} &
  0.00 (0.00) \\ \hline
\begin{tabular}[c]{@{}c@{}}Conv-64 (\textbf{Ours}) (t = 0.75)\end{tabular} &
  \multicolumn{1}{c|}{56.70 (0.69)} &
  \multicolumn{1}{c|}{\textbf{34.13 (0.76)}} &
  \textbf{42.75 (0.74)} &
  \multicolumn{1}{c|}{72.53 (0.56)} &
  \multicolumn{1}{c|}{\textbf{51.33 (0.75)}} &
  \textbf{58.60 (0.70)} \\ \hline 
\begin{tabular}[c]{@{}c@{}}Conv-64 (\textbf{Ours}) (t = 0.70)\end{tabular} &
  \multicolumn{1}{c|}{56.42 (0.70)} &
  \multicolumn{1}{c|}{\textbf{37.20 (0.76)}} &
  \multicolumn{1}{c|}{\textbf{44.88 (0.73)}}
   &
  \multicolumn{1}{c|}{71.96 (0.56)} &
  \multicolumn{1}{c|}{\textbf{54.12 (0.76)}} &
  \multicolumn{1}{c|}{\textbf{60.41 (0.68)}}
   \\ \hline \hline
\begin{tabular}[c]{@{}c@{}}ResNet-12  (Baseline)\end{tabular} &
  \multicolumn{1}{c|}{68.40 (0.71)} &
  \multicolumn{1}{c|}{0.01 (0.01)} &
  0.00 (0.00) &
  \multicolumn{1}{c|}{81.81 (0.52)} &
  \multicolumn{1}{c|}{0.07 (0.02)} &
  0.00 (0.00) \\ \hline
\begin{tabular}[c]{@{}c@{}}ResNet-12 (\textbf{Ours})\end{tabular} &
  \multicolumn{1}{c|}{65.03 (0.72)} &
  \multicolumn{1}{c|}{\textbf{60.55 (0.76)}} &
  \textbf{62.05 (0.74)} &
  \multicolumn{1}{c|}{79.79 (0.55)} &
  \multicolumn{1}{c|}{\textbf{75.93 (0.61)}} &
  \textbf{77.08 (0.59)} \\ \hline \hline
  \begin{tabular}[c]{@{}c@{}}WRN-28-10 (Baseline)\end{tabular} &
  \multicolumn{1}{c|}{69.55 (0.73)} &
  \multicolumn{1}{c|}{0.00 (0.00)} &
  0.00 (0.00) &
  \multicolumn{1}{c|}{81.35 (0.58)} &
  \multicolumn{1}{c|}{0.00 (0.00)} &
  0.00 (0.00) \\ \hline
\begin{tabular}[c]{@{}c@{}}WRN-28-10 (\textbf{Ours})\end{tabular} &
  \multicolumn{1}{c|}{61.16 (0.73)} &
  \multicolumn{1}{c|}{\textbf{55.25 (0.77)}} &
  \multicolumn{1}{c|}{\textbf{56.36 (0.77)}}
   &
  \multicolumn{1}{c|}{75.64 (0.67)} &
  \multicolumn{1}{c|}{\textbf{68.94 (0.78)}} & 
  \multicolumn{1}{c|}{\textbf{69.92 (0.77)}}
   \\ \hline
\end{tabular}
\endgroup
}
\label{tab:table1}
\end{table*}
The clean accuracy on the WRN-$28$-$10$ baseline model observed was slightly improved than the ResNet-$12$ model. We used the same training procedure as which was used for training ResNet-$12$ and did not focus much on hyperparameter optimization as we wanted to showcase the  utility of our method. While training the WRN-$28$-$10$ with our approach, as per trend, we observed some drop in clean accuracy but found an absolute boost in robust accuracy by $\approx55\%$. Also, we observe that this improvement in robust accuracy is slightly lower than that for the Resnet-$12$, as the deeper models are more vulnerable to adversarial attacks \cite{szegedy2013intriguing,goodfellow2014explaining}. As expected, the results obtained from the conv $(64)_{\times4}$ baseline model were less than the Resnet-$12$ baseline model due to its low capacity. Because of its low capacity, it was difficult to achieve the fixed threshold value of $98.0$ during training that was used throughout all the previous experiments for shifting the radius distribution towards low radius. In order to counter this, we reduced the threshold value for conv $(64)_{\times4}$  and repeated the experiment with threshold equal to $75.0$ and $70.0$. The clean accuracy for conv $(64)_{\times4}$  using our approach (trained with reduced threshold) is almost similar to the clean accuracy of conv $(64)_{\times4}$  baseline, which is different from the deeper models WRN-$28$-$10$ and Resnet-$12$ where the performance drop in clean accuracy of our model compared to the baseline was noticeable. The robust accuracy of conv $(64)_{\times4}$ using our approach is enhanced by 35\% when compared to that of the baseline.The difference between robust and clean accuracy of our proposed conv $(64)_{\times4}$ model (row-$2$ and row-$3$ in table-$1$) is $\approx20\%$. This behavior is strikingly different from relatively deeper architectures such as ResNet-$12$ and WRN-$28$-$10$ where the difference is $5\%$ (row-$5$) and 6\% (row-$7$) respectively. 
ResNet-$12$ achieves better results than WideResnet as expected. However, conv $(64)_{\times4}$ being a shallower network has performed poorly than ResNet-$12$. This specific behaviour of shallow networks is an interesting observation that would require further investigation and might be an interesting direction to explore in the future.
\subsection{Finetuning with Cosine Similarity loss ($L_{cs}$) on either support, query or both} 
We train our student model $S$ on low-frequency images at the pretraining and fine-tuning stage using $L_{cs}$ loss so that model performs well during evaluation on these cut-off low-frequency images. Motivated by the transductive finetuning \cite{Dhillon2020A}, which showed a lot of improvement in performance by applying $L_{e}$ on the query samples, we apply $L_{cs}$ loss on the unlabelled query set during fine-tuning. Apart from having $L_{cs}$ on the query set, it could also be applied on the support set or both the support and query set at the fine-tuning stage. So in this section, we study the effect of opting the other choices for $L_{cs}$ regularisation on the performance of model $S$ while evaluating the query set. The results of the study are shown in Table \ref{tab:table2}.

\begin{table}[htp]
\caption{Ablation on different options on which cosine similarity loss ($L_{cs}$) can be applied during finetuning. Frequency regularization on query set obtains better adversarial accuracy as compared to support set. It also yields similar performance as that of combined query and support set at a lesser computational cost.} 
\centering
\scalebox{0.90}{
\begingroup
\setlength{\tabcolsep}{0.18cm} 
\renewcommand{\arraystretch}{1.6}
\begin{tabular}{|l|l|ccc|ccc|}
\hline
\multicolumn{1}{|c|}{Setup} & \multicolumn{1}{c|}{Method} &
  \multicolumn{1}{c|}{Clean} &
  \multicolumn{1}{c|}{PGD} &
  \begin{tabular}[c]{@{}c@{}}Auto Attack\end{tabular} \\ \hline \hline
\multirow{4}{*}{1 - SHOT} & Baseline &
  \multicolumn{1}{c|}{68.40 (0.71)} &
  \multicolumn{1}{c|}{0.01 (0.01)} &
  0.00 (0.00) \\ \cline{2-5} \cline{2-5} 
& \begin{tabular}[c]{@{}l@{}}$L_{cs}$ on support\end{tabular} &
  \multicolumn{1}{c|}{64.18 (0.75)} &
  \multicolumn{1}{c|}{52.79 (0.79)} &
  56.15 (0.77)  \\ \cline{2-5} 
& \begin{tabular}[c]{@{}l@{}}$L_{cs}$ on query\end{tabular} &
  \multicolumn{1}{c|}{65.03 (0.72)} &
  \multicolumn{1}{c|}{60.55 (0.76)} &
  62.05 (0.74)  \\ \cline{2-5} 
& \begin{tabular}[c]{@{}l@{}}$L_{cs}$ on (query + support)\end{tabular} &
  \multicolumn{1}{c|}{65.25 (0.72)} &
  \multicolumn{1}{c|}{61.22 (0.75)} &
  62.74 (0.74)  \\ \hline
\multirow{4}{*}{5 - SHOT} & Baseline &
  \multicolumn{1}{c|}{81.81 (0.52)} &
  \multicolumn{1}{c|}{0.07 (0.02)} &
  0.00 (0.00) \\ \cline{2-5} \cline{2-5} 
& \begin{tabular}[c]{@{}l@{}}$L_{cs}$ on support\end{tabular} & \multicolumn{1}{c|}{78.76 (0.55)} &
  \multicolumn{1}{c|}{68.58 (0.65)} &
  71.09 (0.62) \\ \cline{2-5} 
& \begin{tabular}[c]{@{}l@{}}$L_{cs}$ on query\end{tabular} &
  \multicolumn{1}{c|}{79.79 (0.55)} &
  \multicolumn{1}{c|}{75.93 (0.61)} &
  77.08 (0.59) \\ \cline{2-5} 
& \begin{tabular}[c]{@{}l@{}}$L_{cs}$ on (query + support)\end{tabular} &
  \multicolumn{1}{c|}{80.51 (0.51)} &
  \multicolumn{1}{c|}{77.12 (0.60)} &
  78.16 (0.58) \\ \hline  
\end{tabular}
\endgroup
}
\label{tab:table2}
\end{table}

For regularisation using support set, the clean accuracy was almost similar to that of performance with $L_{cs}$ loss on query set, whereas a significant 
reduction in robust accuracy is observed. As expected, it did not help much due to fewer samples in the support set compared to the unlabelled query set. When low-frequency regularization was applied on both support and query sets, the performance slightly improved compared to the query set. Practically for an $n$-way $k$-shot scenario with a large $n$ and even for small $k$, the total samples in support set, i.e., $n \times k$, would be extensive, and training followed with evaluation on it for a large number of episodes  $(\approx1000)$ would significantly increase the run-time and computational expense. Since the performance margin is not significant we prefer to use only the query set regularization to get a better time efficiency.

\section{Sensitivity analysis over Radius Range ($r^{range}$) in Progressive Learning}
The experiment section V-D of the main paper demonstrated that radius selection is crucial for the robust and clean accuracy tradeoff. Choosing a low radius for frequency cutoff would have removed most of the adversarial perturbation but resulted in low discriminativeness. On the other hand, frequency cutoff at a high radius would have conserved image discriminability but still would have a higher adversarial content. Hence to overcome this, we proposed a progressive way of selecting the radius from list $r^{range}$ (as presented in eq.($6.$)) using a weighted sampling distribution. We considered the  $r^{range}$ to be [$n/2$, $2$] throughout all the experiments. Since radius selection is critical, we do a sensitivity analysis over different choices of $r^{range}$ to check the performance dependency of our approach on this. Apart from [$n/2$,$2$], we took [$n$-$1$, $2$] as $r^{range}$, and performed experiments on the CIFAR-FS dataset with $r^{range}$ equals to [$31$,$2$].  

\begin{table}[htp]
\caption{Investigating the impact of varying the radius range ($r^{range}$) in progressive learning on the model performance. Reducing the $r^{range}$ from [$31$,$2$] to [$16$,$2$] leads to a minor decrease and almost same performance.}
\centering
\scalebox{0.90}{
\begingroup
\setlength{\tabcolsep}{0.18cm} 
\renewcommand{\arraystretch}{1.6}
\begin{tabular}{|c|c|ccc|ccc|}
\hline
\multicolumn{1}{|c|}{Setup} & \multicolumn{1}{c|}{Method} &
  \multicolumn{1}{c|}{Clean} &
  \multicolumn{1}{c|}{PGD} &
  \begin{tabular}[c]{@{}c@{}}Auto Attack\end{tabular} \\ \hline \hline
\multirow{3}{*}{1 - SHOT} & \begin{tabular}[c]{@{}c@{}}PL on r : {[}16,2{]} \\ ({[}n/2, 2{]})\end{tabular} &
  \multicolumn{1}{c|}{65.03 (0.72)} &
  \multicolumn{1}{c|}{60.55 (0.76)} &
  62.05 (0.74) \\ \cline{2-5}
& \begin{tabular}[c]{@{}c@{}}PL on r : {[}31,2{]} \\ ({[}n-1, 2{]})\end{tabular} &
  \multicolumn{1}{c|}{65.60 (0.71)} &
  \multicolumn{1}{c|}{61.87 (0.76)} &
  62.96 (0.73) \\ \hline
  
\multirow{3}{*}{5 - SHOT} & \begin{tabular}[c]{@{}c@{}}PL on r : {[}16,2{]} \\ ({[}n/2, 2{]})\end{tabular} &
  \multicolumn{1}{c|}{79.79 (0.55)} &
  \multicolumn{1}{c|}{75.93 (0.61)} &
  77.08 (0.59) \\ \cline{2-5}
& \begin{tabular}[c]{@{}c@{}}PL on r : {[}31,2{]} \\ ({[}n-1, 2{]})\end{tabular} &
  \multicolumn{1}{c|}{79.93 (0.55)} &
  \multicolumn{1}{c|}{76.68 (0.60)} &
  77.47 (0.59) \\ \hline
\end{tabular}
\endgroup
}
\label{tab:table4}
\end{table}

Table \ref{tab:table4} shows the result for experiments with different $r^{range}$ i.e [$32/2$,$2$] and [$31$,$2$]. It could be observed that the results for both robust and clean accuracy for $r^{range}$ [$31$,$2$] are slightly better than that for [$16$, $2$]. However, this slight performance gap between the two is not much from which we can conclude that the performance of our approach is not much sensitive to the $r^{range}$ that is very different from the case of fixed radius selection where it was very critical to the performance. Since the performance of our proposed approach is very less sensitive to the $r^{range}$, we have chosen it to be minimum, i.e. [$n/2$, $2$] throughout all the experiments (in the main paper) to get a better time efficiency.

\begin{figure*}[htp]
\centering
\centerline{\includegraphics[width=1.1\textwidth]{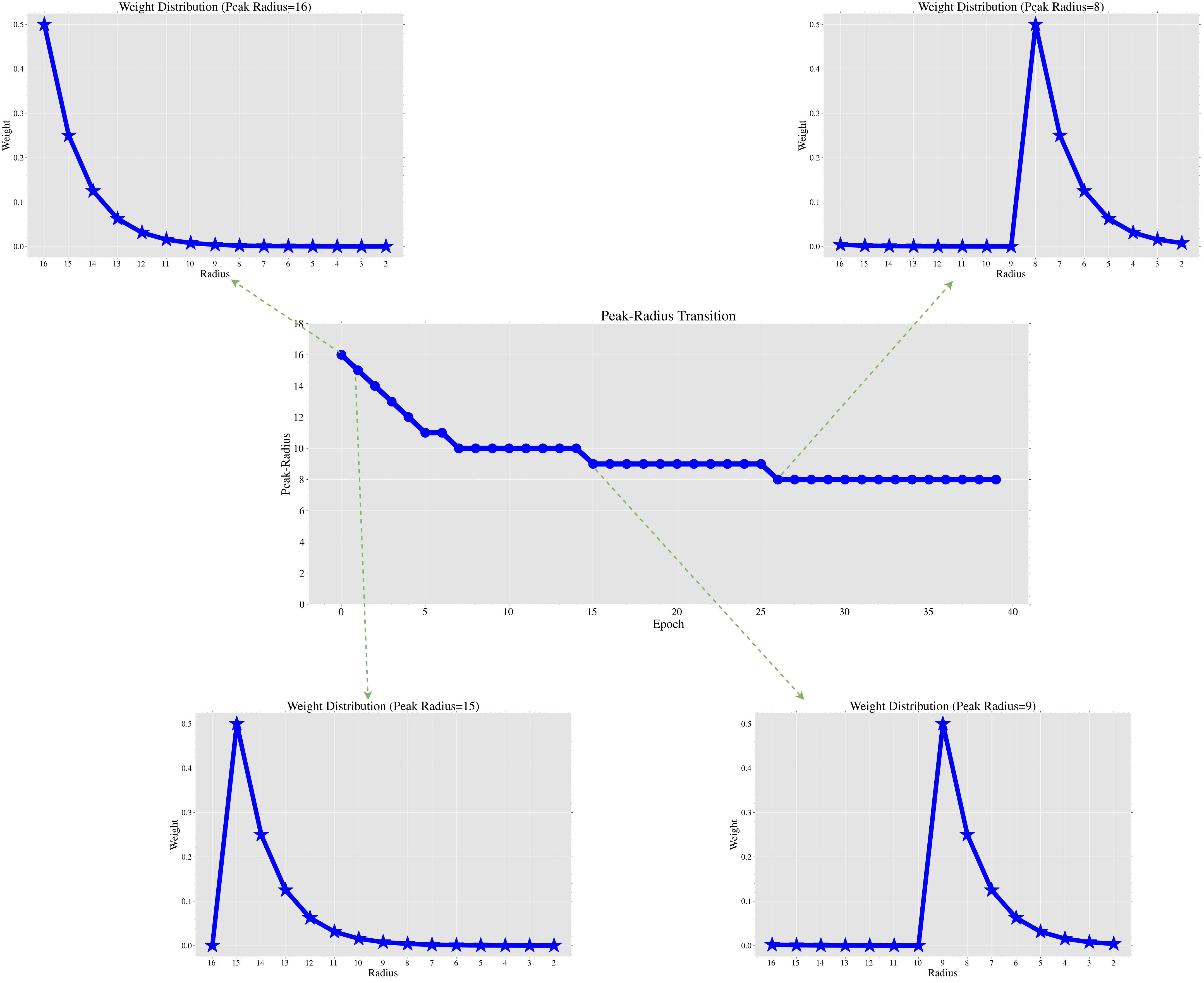}}
\caption{\small{Plot to show the transition in peak radius and its corresponding long tail distribution over radii that are used at each epoch during pretraining stage. Keeping the ease of demonstration in mind, we show the weight-distribution for a few high and a few lower values of peak radius. For each training sample, a radius is sampled from the respective weight distribution to obtain its low-frequency representation. The shift in the distribution (occurs after crossing the training accuracy threshold on low-frequency samples of training data) is more frequent at initial epochs and becomes slower at later epochs of training on CIFAR-FS dataset.}} 
\label{fig:radius-distribution-shift}
\end{figure*}

Using the $r^{range}$ as [$n/2$, $2$], we performed progressive learning during the Pretraining stage. Initially, a long tail distribution over the values of $r^{range}$ is computed using eq. $6.$ (in main draft) where the the radius corresponding to highest weight (peak radius $r_{p}$) is at $n/2$ ($16$ in the case of CIFAR-FS dataset). During the course of training, the value of radius $r$ is sampled from the distribution for each sample. The sampled radius is used to split the frequency representation of the respective sample in order to obtain its low-frequency component (refer eq. $3.$ in the main paper). The model is then trained using the losses described in eq. $5.$ in the paper draft. Once the training accuracy on low-frequency samples at radius $r_{p}$ crosses the threshold then the distribution over radii is shifted by taking the peak radius $r_{p}$ as $n/2-1$. Then the weighted sampling is performed from the current shifted distribution and the model training is resumed thereafter. Again when the training accuracy on the low-frequency samples at the current $r_{p}$ exceeds the threshold, the distribution is shifted with peak radius $r_{p}$ as $n/2-2$. This process is continued till the $r_{p}$ value is equal to $r_{min}$ or the maximum number of epochs is reached. Below, we show the weighted distribution and its peak radius $r_{p}$ that are used at each epoch during training (Fig.~\ref{fig:radius-distribution-shift}). It can be observed that the model finds it easy to learn on high radius and the distribution shift occurs more frequently in the initial course of training. On lower radius, the model finds it more harder to learn on corresponding low-frequency samples and thus takes more epochs to cross the training threshold leading to a slower shift in the distribution.
\vspace{0.2in}
\newpage
\section{Additional Training Details}
In line with the previous works on few-short adversarial robustness \cite{goldblum2020adversarially,liu2021longterm} we perform adversarial attacks using 20-step PGD attack with $\epsilon = 8/255$ and step-size = $2/255$. We additionally also evaluate our technique on the current state-of-the-art adversarial attack i.e. AutoAttack using a $\epsilon$ $= 8/255$. We follow the same experimental protocol as described in \cite{Dhillon2020A} for query performance evaluation. We use 1080Ti 12Gb cards for all our experiments. During pretraining, we use an SGD optimizer with a batch size of $128$ and a base learning rate of $1\text{e-}2$ scheduled with cosine annealing for $40$ epochs. During finetuning, we use Adam optimizer with a fixed learning rate of $5\text{e-}5$ for $25$ epochs. We set $\lambda$ as $0.80$ for both datasets (refer eq. $6.$),  $t$ as $98$ and $90$ for CIFAR-FS and Mini-ImageNet, respectively.
\section{Qualitative Demonstration}
\subsection{Visualization in the Frequency Domain}
\begin{figure}[htp]
\centering
\centerline{\includegraphics[width=\textwidth]{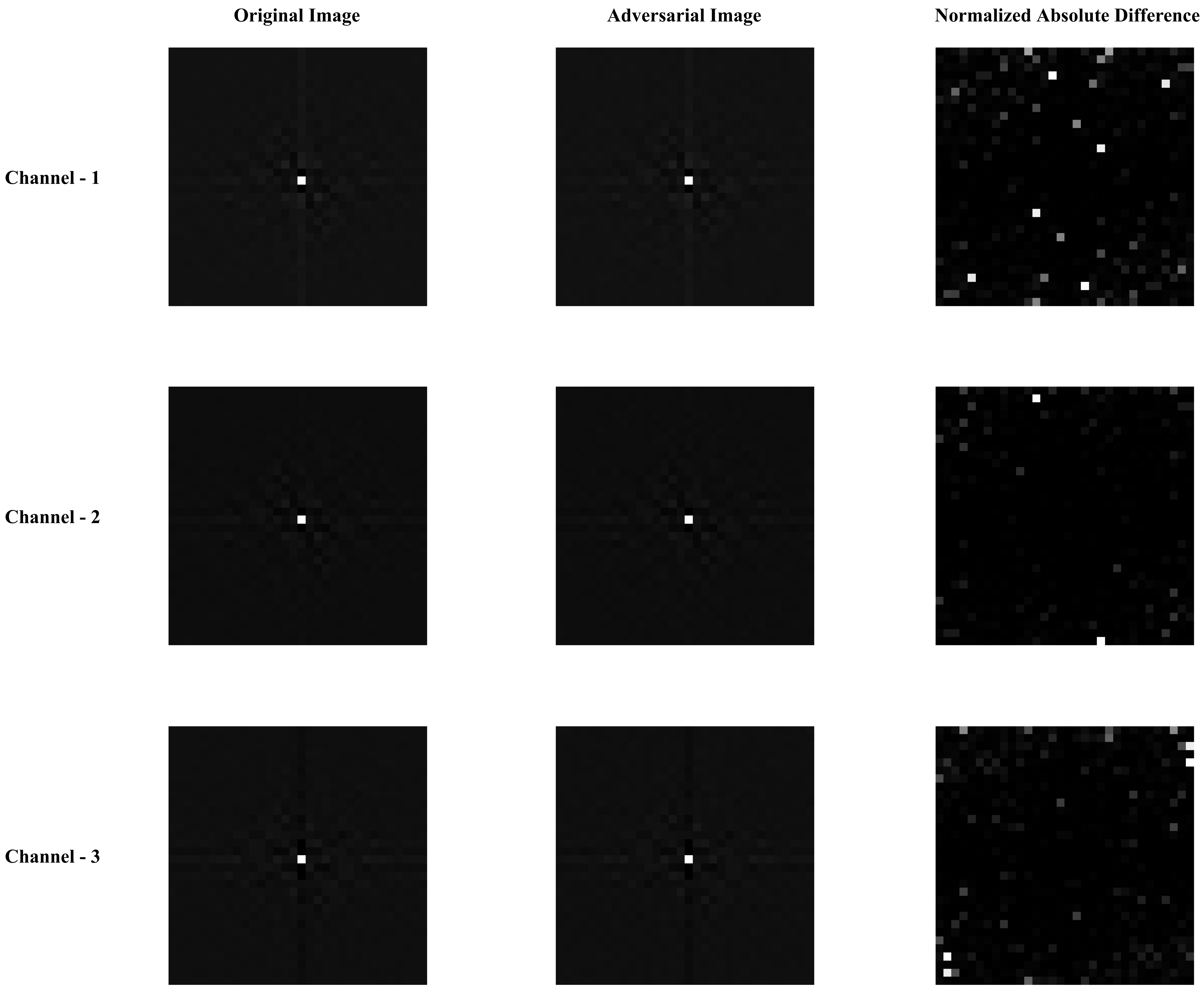}}
\caption{Visualization of the frequency spectrum of a clean image ($FT(x)$), corresponding adversarial image generated ($FT(\tilde{x})$) using PGD adversarial attack and their normalize absolute difference ($\frac{\lvert FT(\tilde{x}) - FT(x)\rvert}{\lvert FT(x)\rvert}$). In the third column, we observe that the absolute normalized difference between Fast Fourier Transform of adversarially perturbed and original image tend to have higher values in high-frequency regions, indicating that adversarial attacks severely contaminates high-frequency regions.
}
\end{figure}
\newpage
\subsection{Visual Demonstration of Proposed Method on Original and Adversarial perturbed Input}
\vspace{0.2in}
\begin{figure*}[htp]
\centering
\centerline{\includegraphics[width=1.12\textwidth]{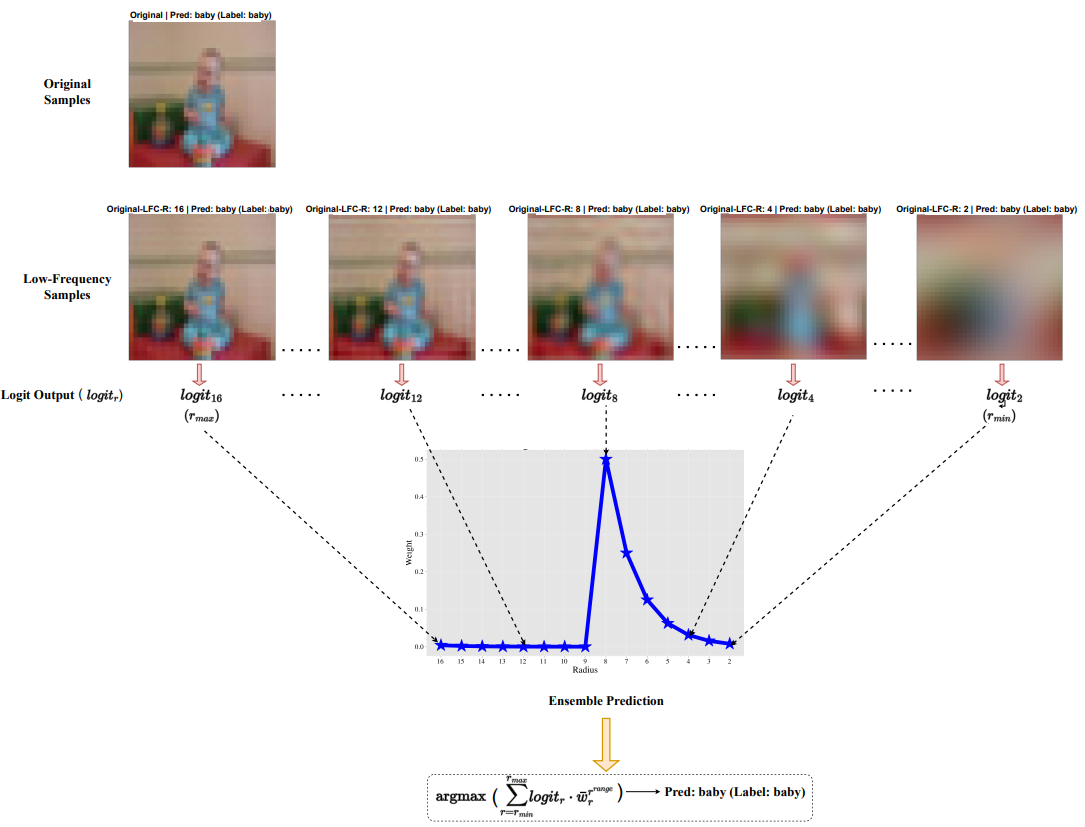}}
\caption{Visualization of Original sample and Low-Frequency sample at different radii. Final ensemble prediction is obtained by the weighted combination of the Logit Outputs corresponding to Low-Frequency samples at different radii.}
\end{figure*}

\vspace{0.2in}
\begin{figure*}[htp]
\centering
\centerline{\includegraphics[width=1.12\textwidth]{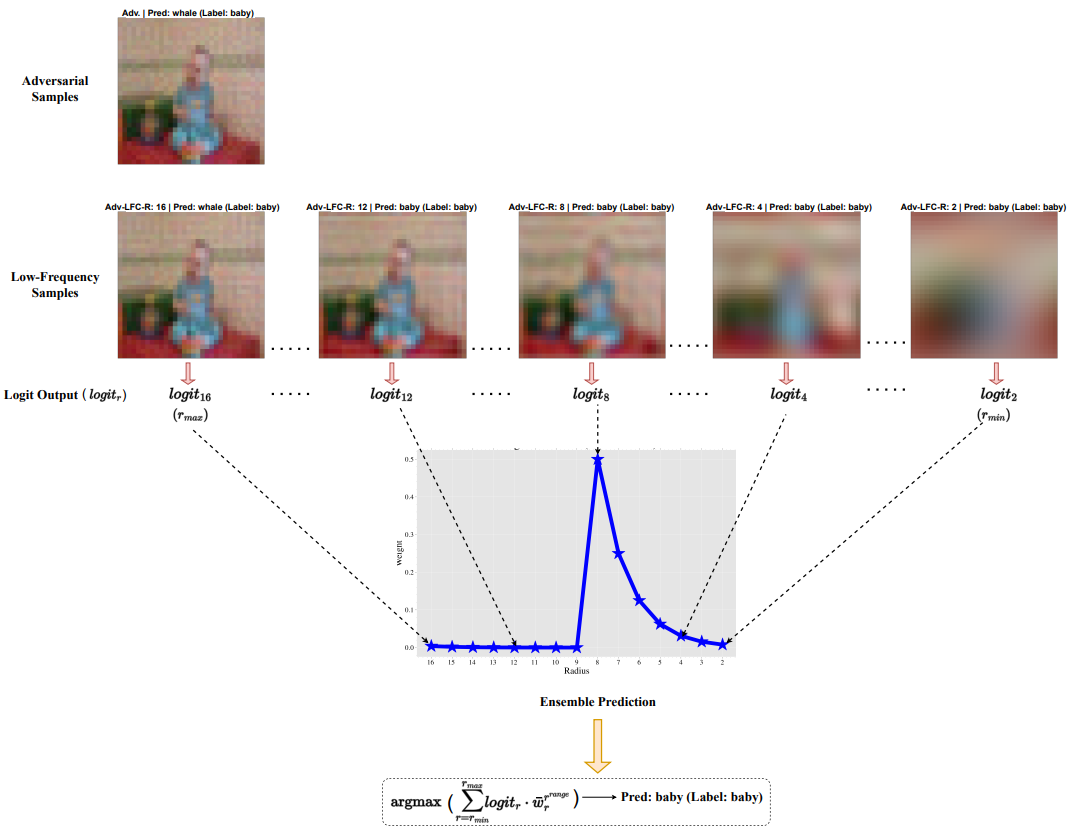}}
\caption{Visualization of Adversarial sample and Low-Frequency sample at different radii. Final ensemble prediction is obtained by the weighted combination of the Logit Outputs corresponding to Low-Frequency Samples at different radii. 
}

\end{figure*}

\newpage
\vspace{0.3in}
\section{Pseudocode of Proposed Algorithm}
\vspace{0.2in}
In this section, we describe the pseudo-code for our proposed approach of robust few-shot learning. Our algorithm is broadly divided into two-stage: a)Pretraining b)Finetuning and Evaluation. In the pretraining stage the network is trained on the base classes, where we follow two distinct steps: a) Teacher training, b)Student training via self-distillation. In the finetuning stage the student model obtained after the pretraining stage is fine-tuned on the novel classes in step-1. The finetuned models are finally evaluated on the unlabeled query set in step-2. The detailed steps of the proposed algorithm are shown in Algorithm~\ref{sup:algo:vendorside}


 \begin{algorithm}
 \caption{Algorithm for our proposed method (DBMA)}
 \label{sup:algo:vendorside}
 \begin{algorithmic}[1]
 \renewcommand{\algorithmicrequire}{\textbf{Input:}}
 \renewcommand{\algorithmicensure}{\textbf{Output:}}
\STATE \textbf{Input:} base class dataset $D_{B}$, novel class dataset $D_{N}$, teacher model $T$, student model $S$
\\ \imp{\underline{\textbf{Stage 1: \textit{\texttt{Pre} training}}}} 
\\ \underline{\textbf{Step 1: \textit{{Teacher} training}}} 
\FOR{$epoch < MaxEpoch$:}
\FOR{$j = 1 : (|D_{B}|/k)$:}
    \STATE \{$x^B_{i}, y^B_{i}\}_{i=1}^{k} \leftarrow$ $j^{th}$ batch of $D_{B}$
    \STATE $L_{ce} \leftarrow$  $\frac{1}{k} \sum_{i=1}^{k} -log (P_{soft}(T(x_{i}^{B}))_{y_{i}^{B}})$
    \STATE \textbf{update} $\theta_{t}$ 
    by minimizing $L_{ce}$ using SGD optimizer
    \ENDFOR
\ENDFOR 
\STATE \underline{\textbf{Step 2: \textit{Student training} via Self-Distillation}} 
\STATE Initialize $\theta_{s}$ (weights of $S$) with $\theta_{t}^{*}$ (weights of $T$ after Step $1$ training).
\STATE $\lambda \leftarrow 0.80$
\STATE $r^{range} \leftarrow [r_{max}, r_{max}-1,..., r_{min}]$
\STATE $w^{r^{range}}_{i} \leftarrow \lambda^{r^{range}_{i}}  \forall i \in [0, r_{max}-r_{min}+1)$
\FOR{$epoch < MaxEpoch$:}
\FOR{$j = 1 : (|D_{B}|/k)$:}
    \STATE \{$x_{i}^{B}, y_{i}^{B}\}_{i=1}^{k} \leftarrow$ $j^{th}$ batch of $D_{B}$
    \STATE $r \leftarrow$ weighted sampling over range [$r_{max}$,$r_{min}$]  
    \STATE $F^{mask_{lr}} \leftarrow$ compute low-frequency mask at $r$
    \STATE $xl_{i}^{B} \leftarrow FT^{-1}(FT(x_{i}) \circ F^{mask_{lr}})$, $\forall i \in [1..k]$
    \STATE $L_{ce} \leftarrow \frac{1}{k} \sum_{i=1}^{k} -log (P_{soft}(S(x_{i}^{B}))_{y_{i}^{B}})$
    \STATE $L_{cs} \leftarrow \frac{1}{k} \sum_{i=1}^{k}  \frac{S(xl_{ir}^{B})^TT(x_{i}^{B})}{ \norm{S(xl_{ir}^{B})} \norm{T(x_{i}^{B})}}$
    \STATE \textbf{update} $\theta_s$ by minimizing $L_{ce}$-$L_{cs}$ using SGD optimizer
    \ENDFOR
\STATE $r_{p} \leftarrow$ radius corresponding to $\mathrm{argmax} (w^{r^{range}})$ 
\STATE $acc_{r_{p}} \leftarrow \frac{1}{b}\sum_{i=1}^{b} \mathds{1} (S(xl_{ir_{p}}^{B}) == y_{i}^{B}$) 
\IF{$acc_{r_{p}} \geq t$} 
\STATE Shift the weighting scheme s.t. $r_{p} \leftarrow r_{p}-1$ 
\ENDIF \hfill \COMMENT{\jnkc{$t$ is threshold}}
\ENDFOR
\\ \imp{\underline{\textbf{Stage 2: \textit{Finetuning and Evaluation}}}} 
\\ \underline{\textbf{Step 1: \textit{{Student} Finetuning}}} 
\STATE $\{D_{N_{s}}$, $D_{N_{q}}\} \leftarrow D_{N}$
\STATE \{$x_{i}^{N_{s}}, y_{i}^{N_{s}}\}_{i=1}^{n_{s}} \leftarrow$ $D_{N_{s}}$
\STATE \{$x_{i}^{N_{q}}\}_{i=1}^{n_{q}} \leftarrow$ $D_{N_{q}}$
\STATE Use weighting scheme obtained at the end of step $2$ in Pretraining stage i.e. $\bar{w}^{r^{range}}$
\FOR{$epoch < MaxEpoch$:}
     
    \STATE $r \leftarrow$ weighted sampling over range [$r_{max}$,$r_{min}$]  
    \STATE $F^{mask_{lr}} \leftarrow$ compute low-frequency mask at $r$
    \STATE $xl_{i}^{N_{q}} \leftarrow FT^{-1}(FT(x_{i}^{N_{q}}) \circ F^{mask_{lr}})$, $\forall i \in [1..n_{q}]$
    \STATE $L_{ce}= 1/n_{s}  \sum_{i=1}^{n_{s}} -log P_{soft}(S(x_{i}^{N_{s}}))_{y_{i}^{N_{s}}}$ 
\STATE $L_{e} = 1/n_{q}  \sum_{i=1}^{n_{q}}   \sum_{j=1}^{k} -P_{soft}(S(x_{i}^{N_{q}}))_{j} log P_{soft}(S(x_{i}^{N_{q}}))_{j} $
\STATE $L_{cs} = 1/n_{q}  \sum_{i=1}^{n_{q}} \frac{S(xl_{ir}^{N_{q}})^T S(x_{i}^{N_{q}})}{\norm{S(xl_{ir}^{N_{q}})} \hspace{0.05in}\norm{S(x_{i}^{N_{q}})} }$
    \STATE \textbf{update} $\theta_{s}$ 
    by minimizing $L_{ce}+L_{e}-L_{cs}$ using Adam optimizer
\ENDFOR 
\\ \underline{\textbf{Step 2: \textit{Evaluation}}} 
\STATE \noindent $\triangleright$ \textit{Student $S$ is evaluated on optimal parameters $\theta_{s}^{'}$ obtained at the end of step $1$} 
\STATE Weighted ensemble using weighting scheme obtained at end of pretraining stage, $\bar{w}^{r^{range}}$) 
\STATE $Pred(x_{i}^{N_{q}}) = \mathrm{argmax}(\sum_{r=r_{min}}^{r_{max}} (S(xl_{ir}^{N_{q}}) \cdot \bar{w}_{r}^{r^{range}}))$
\end{algorithmic}
\end{algorithm}

\end{document}